\documentclass[conference]{IEEEtran}
\IEEEoverridecommandlockouts
\usepackage{cite}
\usepackage{amsmath,amssymb,amsfonts}
\usepackage{algorithmic}
\usepackage{graphicx}
\usepackage{subfigure}
\usepackage{subcaption}
\usepackage{textcomp}
\usepackage{xcolor}
\usepackage{multirow}
\def\BibTeX{{\rm B\kern-.05em{\sc i\kern-.025em b}\kern-.08em
    T\kern-.1667em\lower.7ex\hbox{E}\kern-.125emX}}
\begin{document}

\title{Trajectory Data Mining and Trip Travel Time Prediction on Specific Roads
\\
}

\author{\IEEEauthorblockN{Muhammad Awais Amin}
\IEEEauthorblockA{\textit{Data Science Consultant} \\
\textit{Datamatics Technologies, Islamabad, Pakistan}\\
awais.amin@datamaticstechnologies.com}
\and
\IEEEauthorblockN{Jawad-ur-Rehman Chughtai}
\IEEEauthorblockA{\textit{Department of CS\&IT} \\
\textit{Women University of AJ\&K, Bagh, Pakistan} \\
jawadchughtai@gmail.com}
\and
\IEEEauthorblockN{Waqar Ahmad}
\IEEEauthorblockA{\textit{Senior Design Officer} \\
\textit{Classified institute, Pakistan}\\
wiqiahmad96@gmail.com}
\and

\IEEEauthorblockN{Waqas Haider Bangyal}
\IEEEauthorblockA{\textit{Department of Computer Science} \\
\textit{Kohsar University Murree, Punjab, Pakistan}\\
waqas.bangyal@kum.edu.pk}
\and
\IEEEauthorblockN{Irfan ul Haq}
\IEEEauthorblockA{\textit{Disruptive Innovation Laboratory} \\
\textit{PIEAS, Islamabad, Pakistan}\\
irfanulhaq@pieas.edu.pk}
\and
}

\maketitle

\begin{abstract}
Predicting a trip's travel time is essential for route planning and navigation applications. The majority of research is based on international data that does not apply to Pakistan's road conditions. 
We designed a complete pipeline for mining trajectories from sensors data. 
On this data, we employed state-of-the-art approaches, including a shallow artificial neural network, a deep multi-layered perceptron, and a long-short-term memory, to explore the issue of travel time prediction on frequent routes.
The experimental results demonstrate an average prediction error ranging from 30 seconds to 1.2 minutes on trips lasting 10 minutes to 60 minutes on six most frequent routes in regions of Islamabad, Pakistan.\footnotemark[1] 
\end{abstract}
\footnotetext[1]{979-8-3503-4971-9/24/\$31.00 ©2024 IEEE}

\begin{IEEEkeywords}
Intelligence Transportation Systems, Travel Time Prediction, Artificial Neural Networks, Multi-layer Perceptron, Long-Short Term Memory
\end{IEEEkeywords}

\section{Introduction}
An Intelligent Transport Systems (ITS) refers to the use of smart computing technology, in the transportation to introduce novel services across various areas of transportation. This includes managing modes of transportation and traffic control. ITS aims to enhance user awareness and ensure safer travel experiences. ITS is used in a range of applications, including traffic management tasks, like routing, regulating traffic signals, displaying communication indicators, recognizing license plates, providing intelligent speed measurements, implementing smart surveillance systems, automatically detecting incidents, smart braking systems, identifying vehicles, assisting with parking information and even forecasting weather conditions. These applications can involve real time data and input, from sources to handle complex scenarios. Additionally, with help of historical baseline data, predictive techniques are developed to allow advance modeling and comparison.

Travel Time Prediction (TTP) is a vital part of Intelligent Transportation Systems (ITS), serving as an essential part of navigation applications and Advanced Traveller Information Systems (ATIS). TTP's significance lies in its capacity to offer precise travel time estimates, facilitating efficient trip planning and enhancing the overall user experience. The accuracy of ATIS, fortified by TTP, extends beyond individual convenience, holding profound implications for the operational efficiency of logistics and transportation companies. Through optimized route planning and resource allocation, a dependable ATIS contributes to streamlined operations, diminished congestion, and heightened sustainability. The incorporation of TTP into ITS underscores its status as a fundamental technology, shaping the future of transportation by fostering informed decision-making and effective resource management in day-to-day operations \cite{ref53_haq2022travel}.

Recently, there has been an increase in the use of data-driven techniques, specifically for predicting Travel Time as a regression problem. These approaches use historical data to directly estimate the Travel Time for whole paths or routes, effectively capturing and modeling the intricacies of traffic patterns. There are primarily two types of data driven approaches; methods based on trajectories and methods based on origin destination. Origin Destination based methodologies exclusively prioritize pickup and drop-off locations, disregarding intermediate paths\cite{ref54_wang2019simple}. In contrast, trajectory-based approaches take into account the complete trajectory information, covering the complexities of the entire travel path. While OD-based methods prove efficient in specific scenarios, they may miss vital details embedded in intermediate trajectories crucial for accurate Travel Time predictions. Conversely, trajectory-based approaches showcase a more comprehensive understanding by incorporating the entire travel path, allowing for a nuanced modeling of traffic complexities. This distinction in approach reflects the ongoing evolution and diversification of data-driven strategies in Travel Time prediction. Trajectory-based models, in particular, offer a comprehensive perspective that considers the subtleties of travel patterns, thereby enhancing prediction accuracy \cite{ref55_wang2014travel}. 

In this study, we trained three methodologies: a shallow artificial neural network, a deep neural network (particularly, a multi-layered perceptron), and a recurrent learning model (referred to as long-short-term memory). The models used mined data which collected from sensors to forecast trip time.

The primary contribution of our proposed research is outlined as follows:

\begin{itemize}
    \item Designed a pipeline to mine trajectories to get most frequent routes.
    \item Applied map matching to map all GPS points onto Open Street Map network.
    \item Applied Ramer-Douglas-Peucker algorithm to reduce GPS points while maintaining the shape of GPS trajectory.
    \item Predicted Trip Time on most frequent routes of Islamabad city.
\end{itemize}

The remaining portions of the paper are structured according to the following structure: Section 2 explores related literature. In Section 3, we provide detailed insights into the preprocessing and data preparation procedures. Section 4 presents a thorough overview of the methodology. The experimental results are presented in Section 5, while Section 6 concludes the research.

\section{Related Work}


ITS has the history of over 45 years through the development stages \cite{ref02_ahmed1979analysis}. The most of earlier research on travel time prediction has been carried out using statistical approaches. Subsequently, data driven approaches become the hot cake of this field with plethora of algorithmic specifications. The recent trend emphasis on the empirical computational intelligence approaches, such as Machine Learning, Neural Networks, Bayesian Networks, Fuzzy logic, Deep Learning, and Evolutionary techniques are considered necessary, particularly due to the demonstrated shortcomings and weaknesses of classical approaches in challenging traffic conditions, intricate road settings, and when working with extensive datasets containing both structured and unstructured data.


There are number of techniques for predicting trip time that might be  divided into two main categories, parametric and non-parametric methodologies, parametric  usually assume that relationship with travel time and factors that influence it, examples are linear regression, Kalman filter and ARIMA.
In 1997, the Kalman filter was introduced for multivariate data \cite{ref16_chen2001dynamic}. Chien et al \cite{ref17_chien2003dynamic} made use of data from roadside terminals and implemented the Kalman Filter model to forecast travel time and Guin et al \cite{ref48_guin2006travel} employed the Box-Jenkins method for applying the ARIMA series of models.

Non parametric approaches utilise the historical data learn the relationship of travel time and other factors. ANN, SVR and KNN regression models are examples. KNN utilised and  evaluated using Wilcoxon signed-rank test \cite{ref09_smith1997traffic}. Smith et al \cite{ref10_smith2000parametric} implemented regression techniques with k-nearest neighbour model with adjusted and weighted output components.

Recent advances in programming and application of intelligent transport system (ITS) applications contributed, organization, and management of live data on large-scale transportation networks. This has opened up opportunities for using non-parametric methods in traffic forecasting. Usually, these models rely on data to make predictions so, the accuracy of their findings relies on the quality level of the available data.

The primary goal is to detect clusters of data patterns that closely resemble the current traffic conditions within a specified prediction timeframe. A comparative analysis was undertaken in 1996, examining KNN regression, NN, the historical averaging, and the ARIMA. The findings revealed that K nearest neighbour regression outperformed the other models in terms of its capacity to transfer-ability and its robustness when applied to different data sets. \cite{ref19_smith1996multiple}. 

Additionally, Amirian et al \cite{ref20_amirian2016predictive} used probe vehicles and Geographic Information System (GIS) technologies to calculate travel duration. Hidden Markov Models and other classical machine learning techniques have been presented to forecast the temporal progression of traffic conditions. \cite{ref21_wang2019simple}. Gradient Boosting, Random Forest, Linear Regression were implied in different research \cite{ref22_rupnik2015travel, refx1_bangyal2013recognition, refx2_bangyal2013analysis ,ref23_zhang2015gradient, ref24_moreira2016online}.

Artificial neural networks (ANNs) are mathematical models driven by data and inspired by artificial intelligence. They possess remarkable abilities in classifying and recognising patterns.\cite{ref25_clark1993use}. Artificial neural networks (ANNs) showed significant efficacy, in a variety of transportation applications., such as analysing driver behaviour, developing autonomous vehicles, estimating parameters, maintaining road surfaces, detecting and classifying vehicles, optimising freight operations, predicting traffic patterns, formulating transportation strategies and economic models, managing air and water transportation, operating submersible vehicles, overseeing metro services and controlling traffic. Because of their overall performance and durability, neural networks are capable of providing accurate predictions for traffic datasets. They can generate multi-step forecasts with less effort, are suitable for spatial and temporal datasets, and are capable of handling nonlinear relationships in multivariate contexts \cite{ref26_zhang1998forecasting}.

The WDR model is an amalgamation of the Wide and Deep models with recurrent models, the LSTM model is employed to collect the contextual information of the route, as described in the study by Wang et al. (2014) \cite{ref30_wang2014travel}. While recurrent structures provide the ability to learn the relationships between different parts of a route, they have prohibitively high computation costs for enormous navigation solutions. Using a mix of GRUs and graph convolutional networks (GCNs), the temporal graph convolutional network (T-GCN) was proposed as a solution to address the issue of temporal and spatial interdependence in travel time prediction\cite{ref35_zhao2019t}. 


Performing travel time prediction  on the whole categorised into two categories:

\textbf{Origin-Destination based method:} The travel time information for all OD pairs is required to support the routing decisions. OD-based methods are simpler and more computationally efficient, making them suitable for real-time applications. Young et al \cite{ref49_yang2017origin} tried to explain the reliability of this method. Scientists used this method to predict travel time on highways \cite{ref50_oh2018short}. Lie et al  \cite{ref51_liu2021travel} used open data to and  chu et al \cite{ref52_chu2019deep}
 utilised this technique to make real time travel time prediction and demand of vehicle.

\textbf{Trajectory-based method:} Trajectory based method provide more detailed travel time estimates along specific routes but can be computationally expensive and may require real-time traffic data. Trajectory-based approaches \cite{ref33_yang2005study, ref34_zhang2018deeptravel} consider a route as a unified entity and directly predict its duration. They primarily use recurring structures to represent the route's contextual information (e.g., signals, junctions and turns). Trajectory based methods offer a level of efficiency when it comes to measuring the time it takes to traverse each intersection along a given route as long as there are sufficient complete path.

\section{Data Mining and Preparation}
We have an indigenous dataset of Islamabad City from April to October 2019 that contains hundreds of million entries. Each record contains the vehicle's id, timestamp, latitude, longitude, speed, and altitude, as well as the reason for the GPS timestamp recording. Chughtai et al. and Haq et al. utilised a different subset of the same dataset\cite{ref57_chughtai2022attention, ref56_chughtai2022heterogeneous, ref53_haq2022travel}. 

\subsection{Trajectory Mining}\label{AA}
We have a raw data from GPS trackers. For mining trajectories, we converted all records into trips based on the reason "Ignition On" and "Ignition Off" including all GPS points. We have obtained around 0.85 million trips within the specified months. 

\subsection{Data Reprocessing and Map Matching}\label{AA}
\begin{figure}[htbp]
\centerline{\includegraphics[width=8cm]{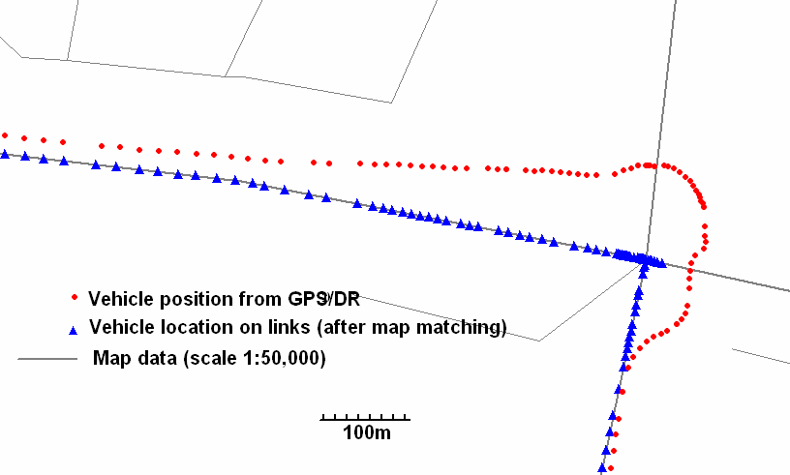}}
\caption{Map Matching}
\label{fig2}
\end{figure}

According to researches, GPS senors is not always accurate \cite{ref44_chao2020survey}, Usually GPS have two kind of errors\cite{ref45_hendawi2020noise}: first measurement errors \cite{ref46_plaudis2021algorithmic} and second is sampling error. Upon visualizing trajectories on map we also have difference between exact location and GPS points visualised. To address issues (as shown in Figure \ref{fig2}), we have employed map-matching techniques.

We have employed an open street map (OSM) for map matching, usually used by most navigation services and mapping service providers. We have added road features like bearing angle and nearest nodes based on GPS points. After setting up road features with GPS points, we utilized the state-of-the-art technique of the Open Street Routing Machine (OSRM) \cite{ref59_shamshad2020parallelized} server to find the nearest nodes against every GPS point using a bearing angle. After assigning the nodes, to visualize the trajectories on the map, we converted the nodes into GPS points again using OSRM. So, our GPS points are a more accurate and correct representation of trajectories after map-matching. 

The final dataset we have compiled consists of all trip-related necessary information. This information includes vehicle identification (id of vehicle), details about the trip itself such as the distance traveled and the average speed, and geographical coordinates for both the starting and ending nodes of the trip. Additionally, there is a representation of the route in form of poly-line, measurements regarding the duration of the trip, and temporal information such as the week's day number, month's day number, and month of the year.
Than filtered out trajectories of busiest routes of Islamabad from dataset based on number of trips on every road and verified as well with local published articles by \cite{ref36_Mazher_2021, ref37_Brothers_2021,  ref38_Brothers_2021}.

\begin{figure}[!htb]
    \centering
    \subfigure[Normal Trajectory]{%
        \includegraphics[width=0.48\linewidth]{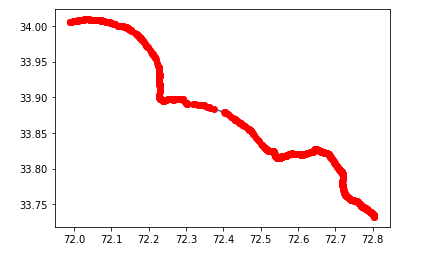}
        \label{fig03_subfigure1}
    }
    \hfill
    \subfigure[Optimal Trajectory]{%
        \includegraphics[width=0.48\linewidth]{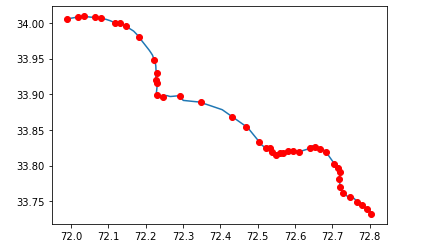}
        \label{fig03_subfigure2}
    }
    \caption{861 points reduced to 46 points only}
    \label{fig03}
\end{figure}

We applied Romar Douglas Peucker algorithm \cite{ref42_saalfeld1999topologically, ref43_zhao2018method} for simplifying trajectories to reduce amount of computation when we have more points they takes more time to assigning nodes, converting back into gps points and also for experimental processing. In our dataset, GPS points recorded based on the action, like turn, brakes, increase or decrease in speed or any action performed by driver. So there were lot of points even for short trips so we need simplified trajectories to utilise economical computations for better results with saving time.  Example of trajectory before and after simplification is shown in Figure \ref{fig03}.

\begin{figure}[htbp]
\centering
\centerline{\includegraphics[width=8cm]{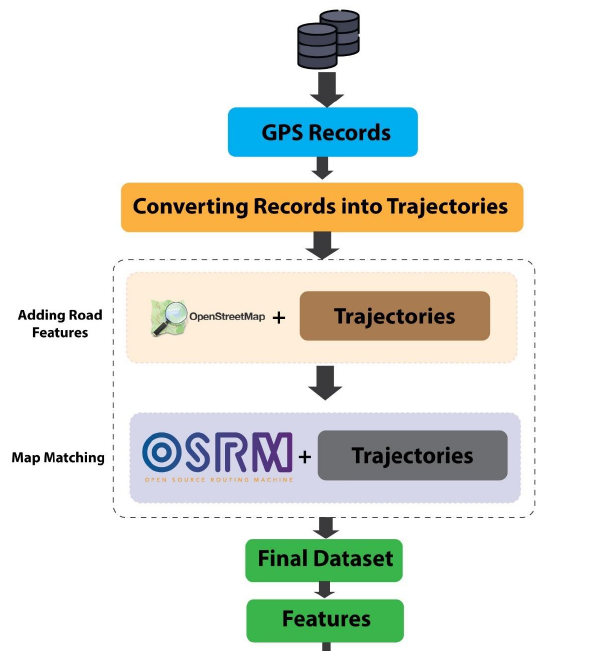}}
\caption{Complete Data Processing.}
\label{fig4}
\end{figure}

Overall, process presented into Figure \ref{fig4}.

\subsection{Data Visualisation}
Finally, from 0.85 million trajectories, we shortlisted 258590 most frequent roads trips within the specified months, including 55215, 47810, 44043, 40199, 36943, and 34380 trips on the Islamabad Expressway, Srinagar Highway, Khayaban-e-Iqbal Road, Jinnah Avenue Road, Faisal Avenue Road, and Agha Shahi Road, respectively. 

\begin{figure}[!htb]
    \centering
    \subfigure[Number of Trips]{%
        \includegraphics[width=0.48\linewidth]{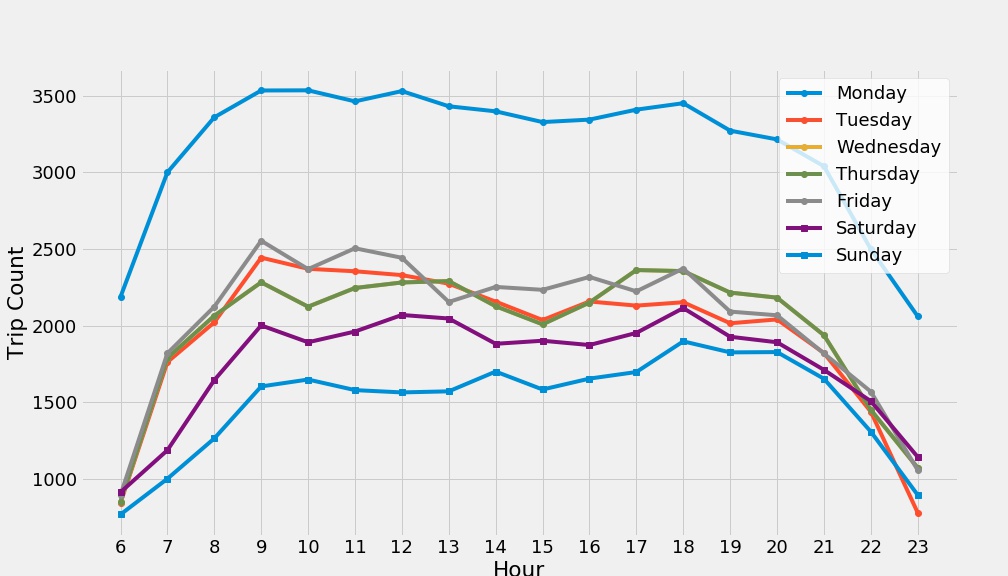}
        \label{fig06_subfigure1}
    }
    \hfill
    \subfigure[Distance of Trips]{%
        \includegraphics[width=0.48\linewidth]{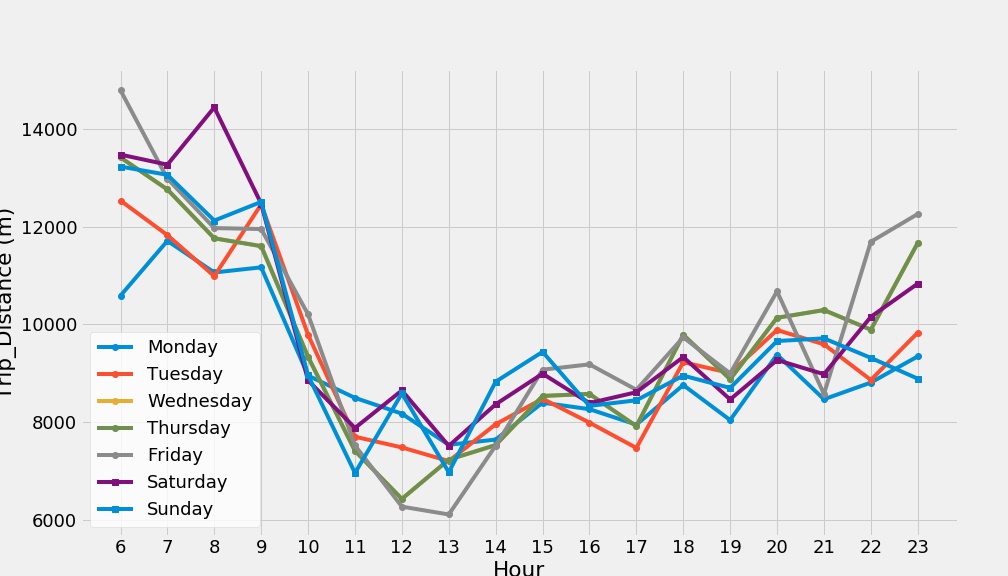}
        \label{fig06_subfigure2}
    }
    \subfigure[Duration of Trips]{%
        \includegraphics[width=0.48\linewidth]{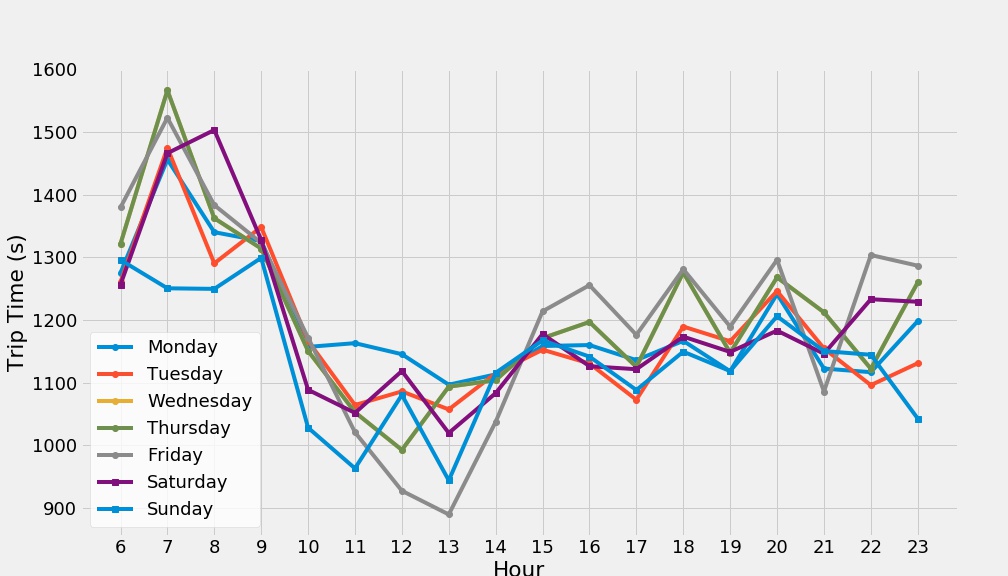}
        \label{fig06_subfigure3}
    }
    \hfill
    \subfigure[Average Speed (KM/H)]{%
        \includegraphics[width=0.48\linewidth]{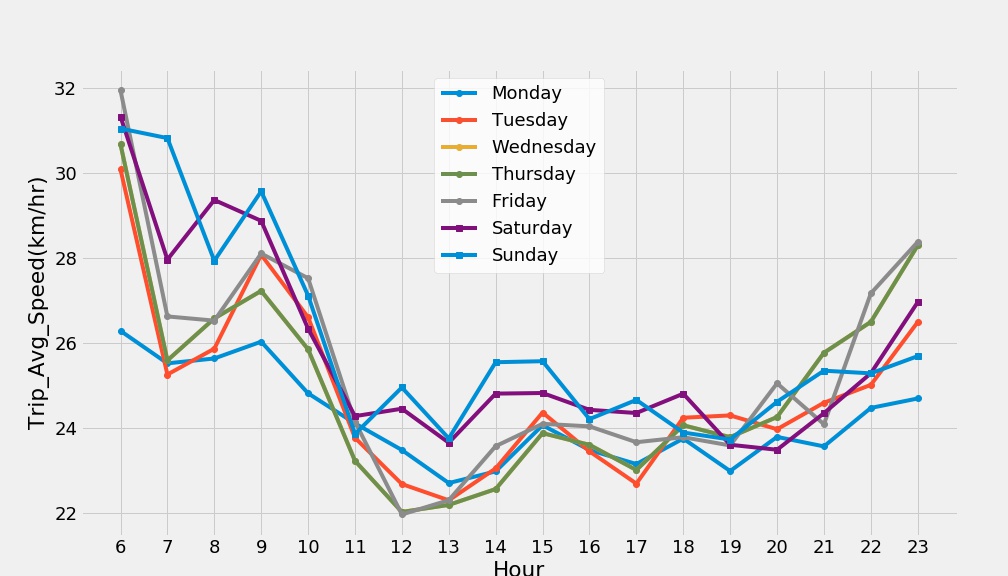}
        \label{fig06_subfigure4}
    }
    \caption{Visualisation of overall data}
    \label{fig06}
\end{figure}

We used data exclusively between 6 a.m. and 11 p.m. to avoid abnormal excursions as highlighted in \cite{ref56_chughtai2022heterogeneous}. In addition to that, we excluded trips lasting more than 60 minutes on individual roads. Figure \ref{fig06} provides some insights of data which we used in this study. Figure \ref{fig06_subfigure1}, \ref{fig06_subfigure2}, \ref{fig06_subfigure3} and \ref{fig06_subfigure4}  illustrate the number of trips, distance of trips, duration of trips and average speed of trips respectively within the specified time frame on various weekdays.  


\section{Methodologies}
\subsection{Artificial Neural Networks}
An ANN design to represent the neurons of human brain similarly like human ANN also have input, nodes, weights and output. It also  called shallow neural networks with one layer.
ANNs alternatively referred to as neural networks. \cite{ref39_krogh2008artificial, refx4_bangyal2023improved}.

\begin{equation}
    \text{Output} = f\left(k + \sum_{i=1}^{z} t_i u_i\right)
    \label{eq:ann}
\end{equation}
Where:
\begin{align*}
    \text{Output} & : \text{Output of the neuron} \\
    f(\cdot) & : \text{Activation function} \\
    k & : \text{Bias term} \\
    t_i & : \text{Input} \\
    u_i & : \text{Weight associated with the input} \\
    z & : \text{Number of inputs}
\end{align*}

\begin{figure}[htbp]
\centering

\centerline{\includegraphics[width=7cm]{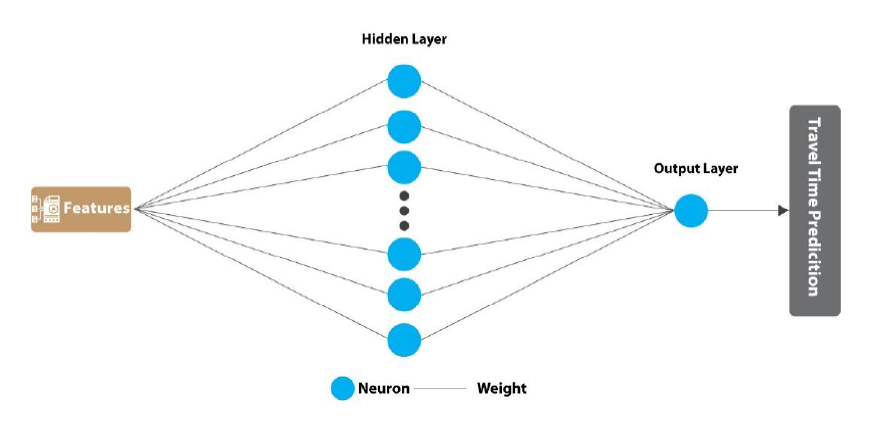}}
\caption{ANN used in experiments}
\label{fig7}
\end{figure}

Equation 1
represents the output of a neuron in an Artificial Neural Network (ANN) by applying the activation function \(f\) to the weighted sum of inputs (\(x_i\) weighted by \(w_i\) and summed up with the bias term \(b\)). We constructed an ANN with a layer of 256 neurons and fed it 12 features as input. The model was trained for 200 epochs using a batch size of 128 as shown in Figure \ref{fig7}.

\subsection{Multi-layer Perceptron}
A MLP is advanced version of ANN with two or more layers. It belongs to the category of neural networks with a mapping between the inputs and outputs. A multi-layer perceptron is made up of an input layer, an output layer, and few hidden layers composed of several stacked neurons. \cite{ref40_popescu2009multilayer, refx3_bangyal2021detection, refx5_bangyal2021analysis}. 
In contrast to Perceptron neurons, which are restricted to using activation functions that enforce a threshold, such as ReLU or sigmoid, neurons in a Multi-layer Perceptron have the flexibility to employ any activation function.

Equation of a two-layer Multilayer Perceptron (MLP):
\begin{equation}
    \text{Output} = f\left(\mathbf{U}^{(3)} \cdot f\left(\mathbf{U}^{(2)} \cdot f\left(\mathbf{U}^{(1)} \cdot \mathbf{x} + \mathbf{k}^{(1)}\right) + \mathbf{k}^{(2)}\right) + \mathbf{k}^{(3)}\right)
\end{equation}


Where:
\begin{align*}
    \text{Output} & : \text{Final output of the MLP} \\
    f(\cdot) & : \text{Activation function} \\
    \mathbf{x} & : \text{Input vector} \\
    \mathbf{U}^{(1)}, \mathbf{U}^{(2)}, \mathbf{U}^{(3)} & : \text{Weight matrices} \\
    \mathbf{k}^{(1)}, \mathbf{k}^{(2)}, \mathbf{k}^{(3)} & : \text{Bias vectors}
\end{align*}

\begin{figure}[htbp]
\centering

\centerline{\includegraphics[width=7cm]{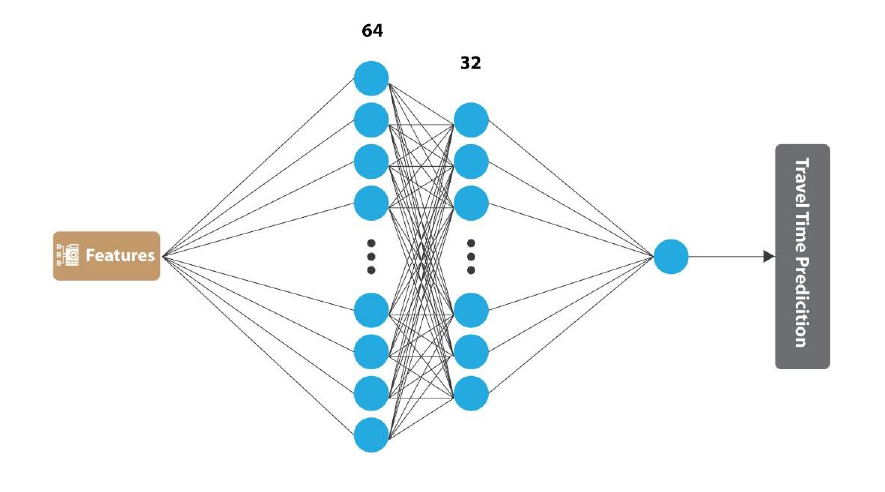}}
\caption{MLP used in experiments}
\label{fig8}
\end{figure}

Equation 2
represents the equation of a two-layer Multilayer Perceptron (MLP), which contains 64 and 32 neurons in layers, with relu serving as the activation function. The model was trained for 200 epochs using a batch size of 128 as shown in figure \ref{fig8}.

\subsection{Long short-term memory (LSTM)}

\begin{figure}[htbp]
\centering

\centerline{\includegraphics[width=6cm]{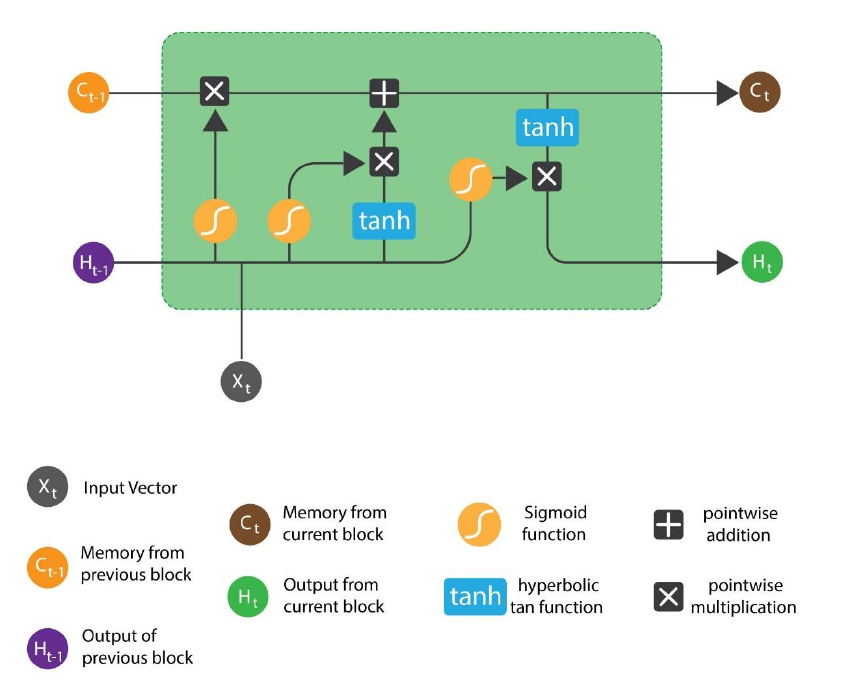}}
\caption{Representation of LSTM Cell}
\label{fig9}
\end{figure}

LSTM is a distinct type of RNN that incorporates the output from the previous stage into the current step. It addresses the issue of long-term dependencies of information stored in the long-term memory while also providing more precise predictions based on prior knowledge. LSTMs designed specifically to handle sequential data as time series, speach and text.The ability to acquire and retain knowledge about complex relationships over extended periods of time in sequential data which make well suited for above mentioned tasks. (Figure \ref{fig9})

The equations for gates in a recurrent neural network (RNN) are given by:

\begin{figure}[htbp]
  \begin{minipage}{0.5\textwidth} 
    \includegraphics[width=6cm]{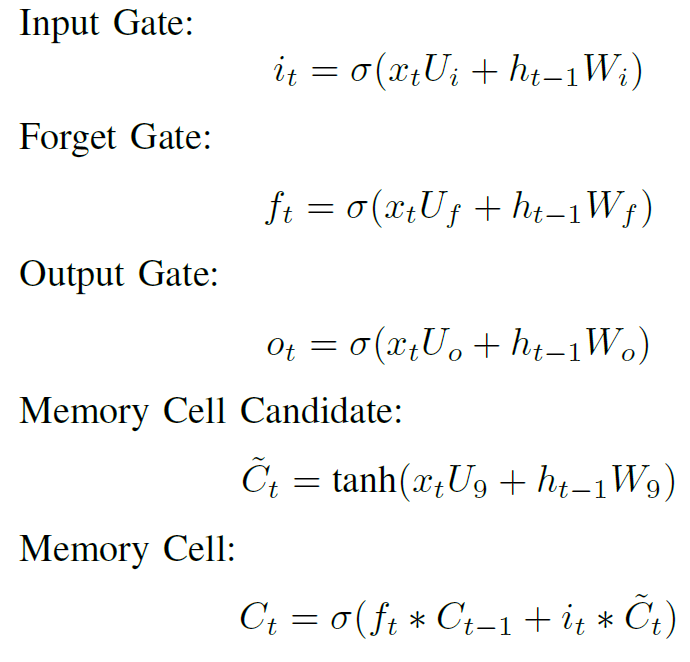}
    \label{fig:lstm}
  \end{minipage}%
\end{figure}












\newpage
\begin{figure}[htbp]
  \begin{minipage}{0.5\textwidth} 
    \includegraphics[width=6cm, height=1cm]{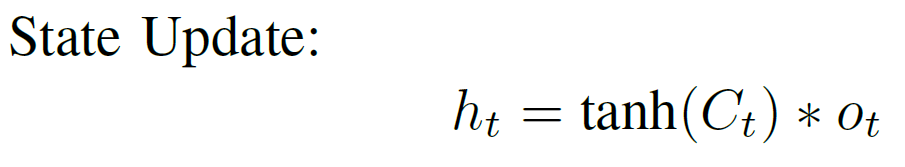}
    \label{fig:lstm}
  \end{minipage}%
\end{figure}

Where:
\begin{align*}
i_t, f_t, o_t & : \text{Input, Forget, Output Gates} \\
\tilde{C}_t & : \text{Memory Cell Candidate} \\
C_t & : \text{Memory Cell} \\
h_t & : \text{State} \\
\sigma(\cdot) & : \text{Activation function} \\
\text{tanh}(\cdot) & : \text{Tanh activation function}
\end{align*}



The Forget gate, Candidate gate , Input gate , and Output gate  are neural networks with a single layer and utilise the Sigmoid activation function. Nevertheless, the Candidate gate employs the Tanh activation function. The gates initially receive the input vector \(x_t \cdot U\) and the prior hidden state \(h_{t-1} \cdot W\), combine them, and subsequently employ the activation function\cite{ref41_zhang2015top}

\begin{figure}[htbp]
\centering

\centerline{\includegraphics[width=9cm]{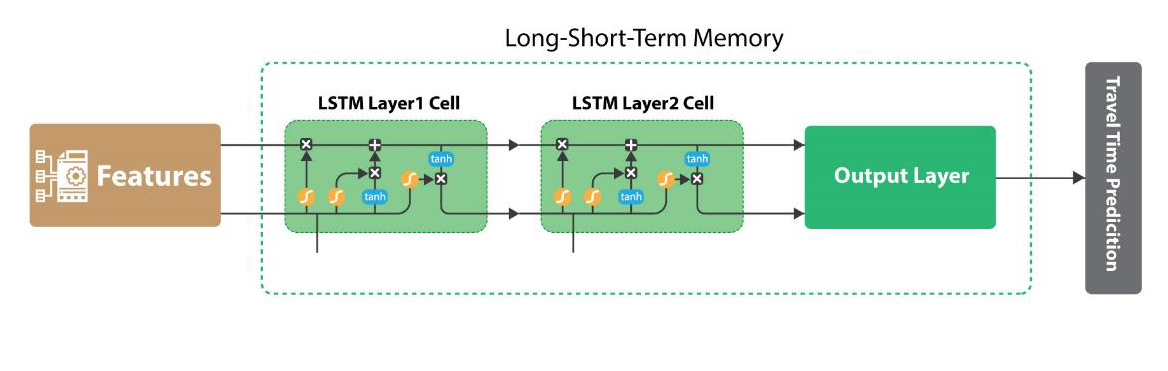}}
\caption{LSTM used in experiments}
\label{fig10}
\end{figure}

We utilized two LSTM layers, each consisting of 64 neurons, and employed the ReLU activation function as shown in Figure \ref{fig10}.

\section{Results \& Discussion}
We collected road data based on trip availability and separated it into subsets of roads. After performing certain fundamental operations of EDA on each dataset’s subset, after ensuring the data's smoothness, we partitioned it into a training set and a validation set. For our experiments, we used five months of data: April, May, June, July, and August for model training, then September and October for testing/validation. After cleaning and ensuring smoothness, we moved towards experiments. We used 12 features in this study which are \textit{Vehicle ID, Trip Distance, Trip Avg Speed(km/hr), Start Lat, Start Lon, End Lat, End Lon, Minutes , Seconds, Day of week, Day of month,  and Month of year} We have employed the chosen models after assessing the characteristics of the data and conducting conclusive experiments on the selected features.

\begin{figure}[!htb]
    \centering
    \subfigure[Islamabad Expressway]{%
        \includegraphics[width=0.48\linewidth]{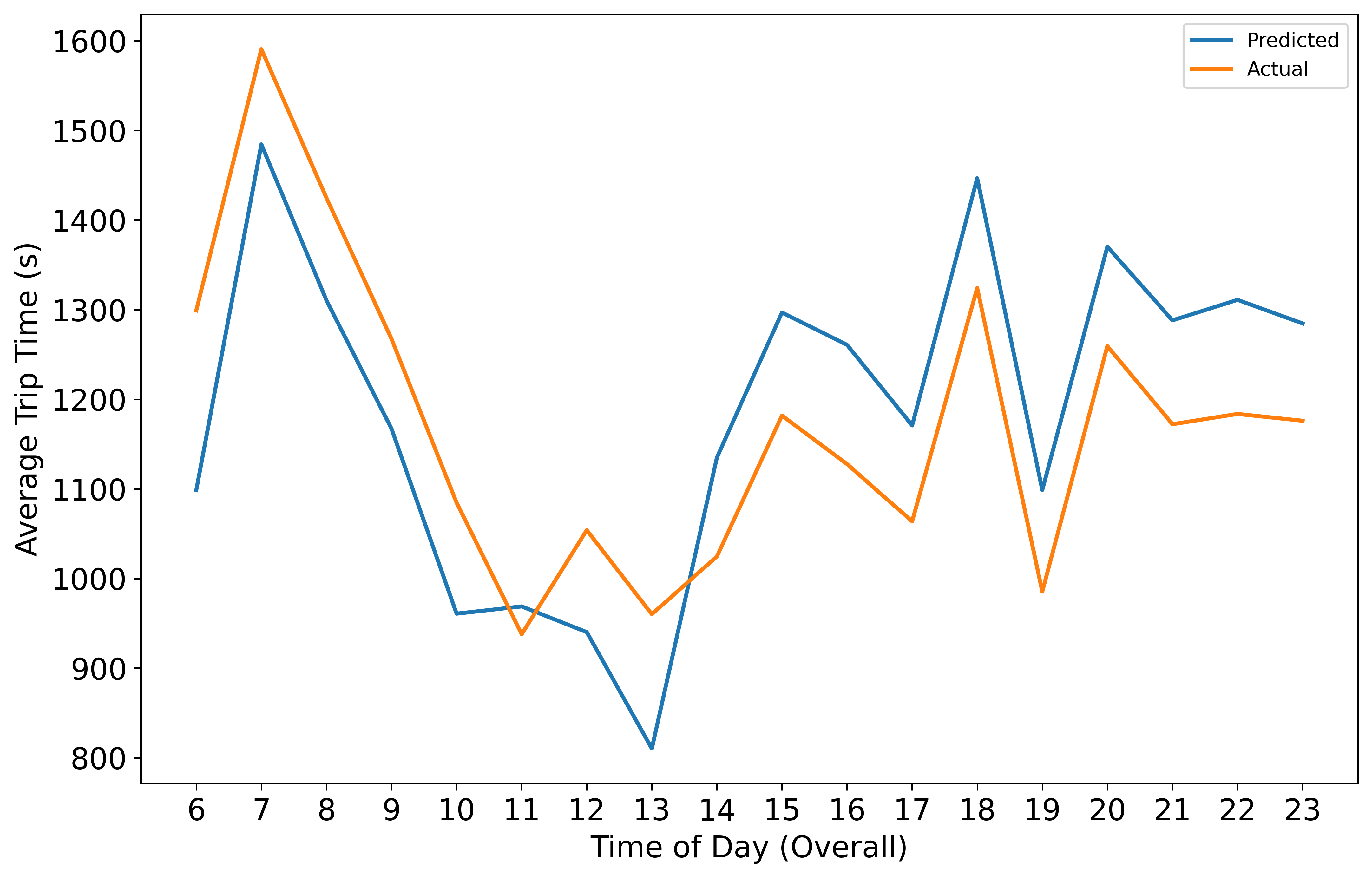}
        \label{fig11_subfigure1}
    }
    \hfill
    \subfigure[Srinagar Highway]{%
        \includegraphics[width=0.48\linewidth]{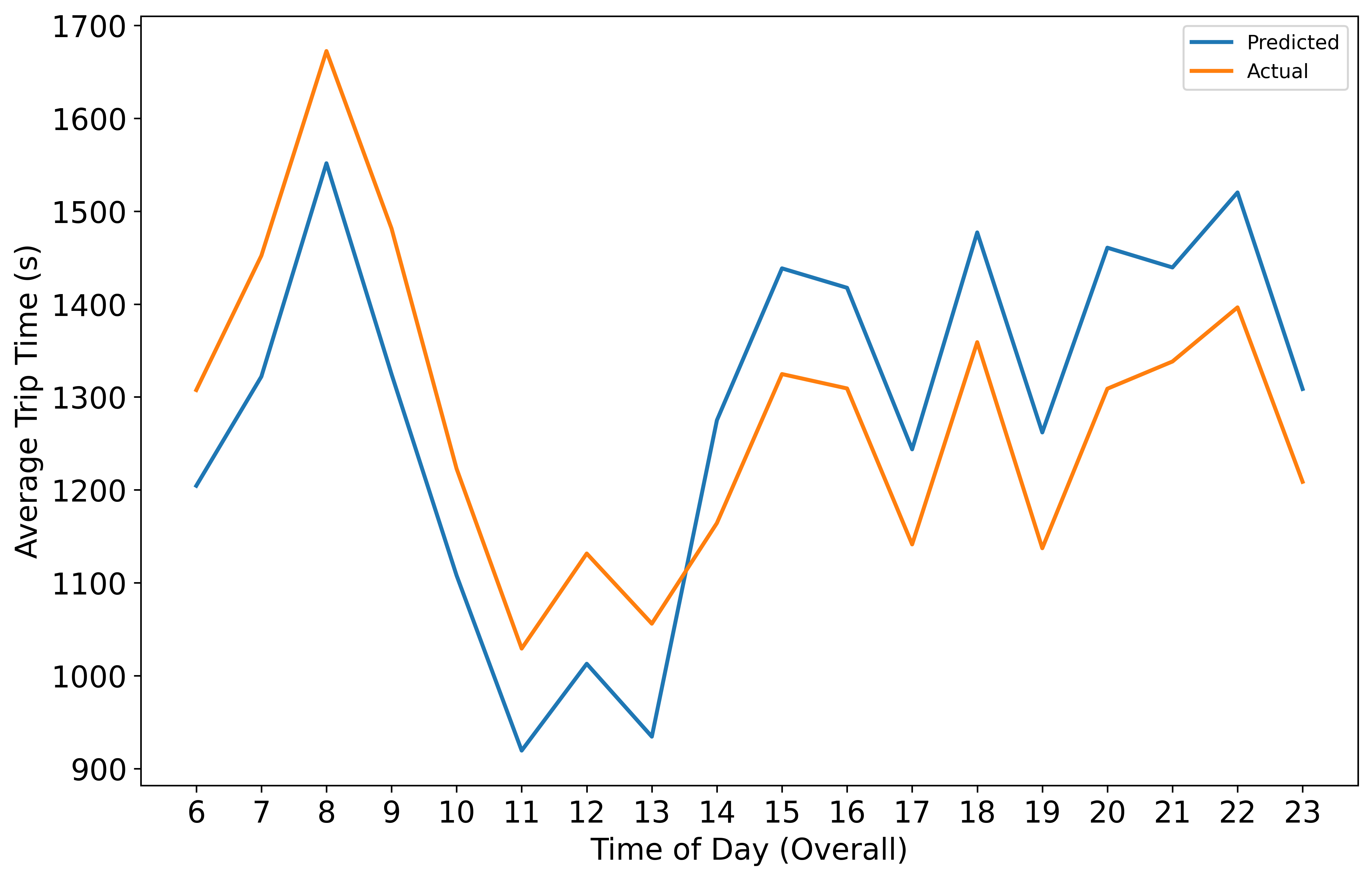}
        \label{fig11_subfigure2}
    }
    \subfigure[Khayaban-e-Iqbal]{%
        \includegraphics[width=0.48\linewidth]{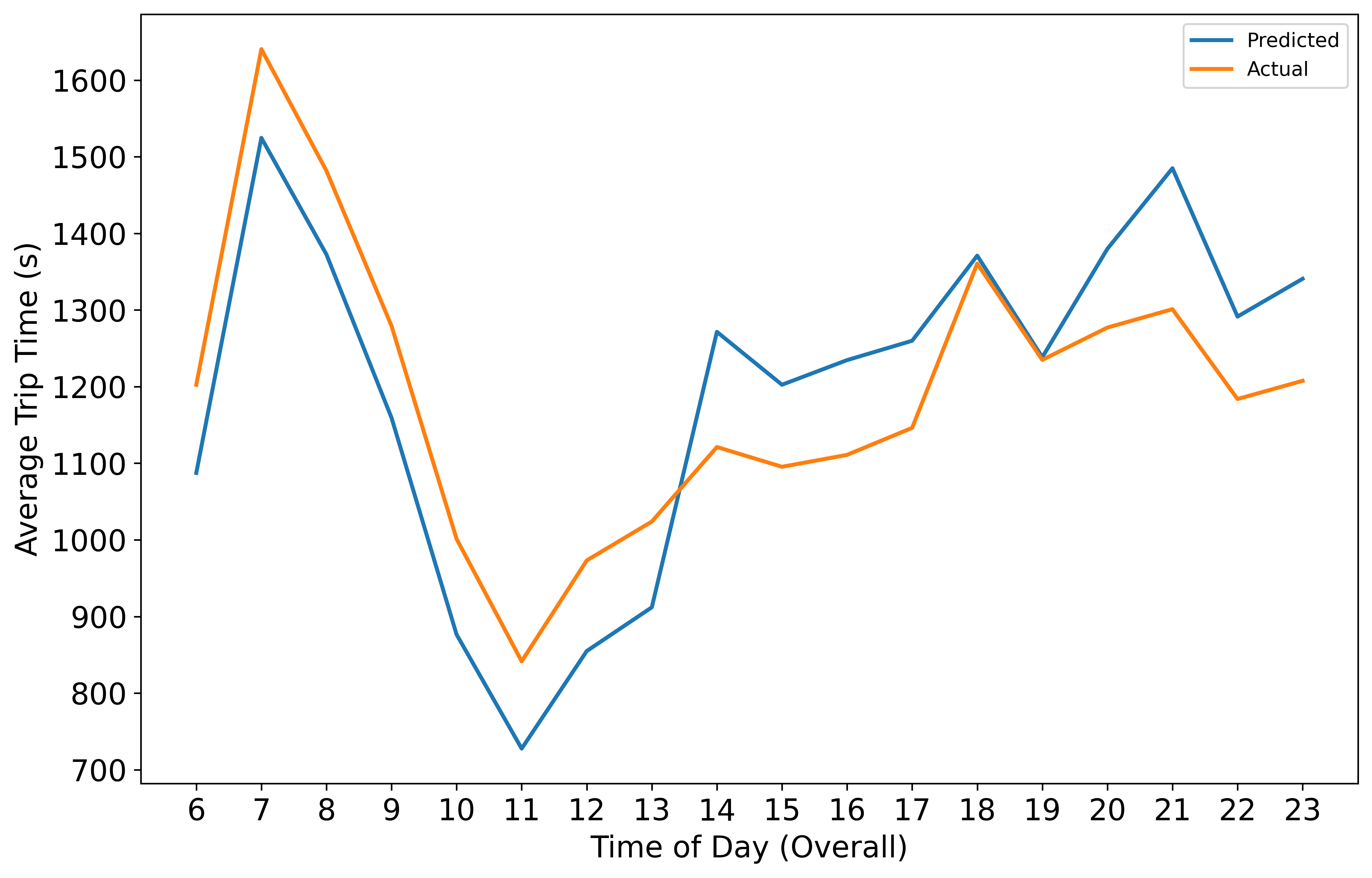}
        \label{fig11_subfigure3}
    }
    \hfill
    \subfigure[Jinnah Avenue]{%
        \includegraphics[width=0.48\linewidth]{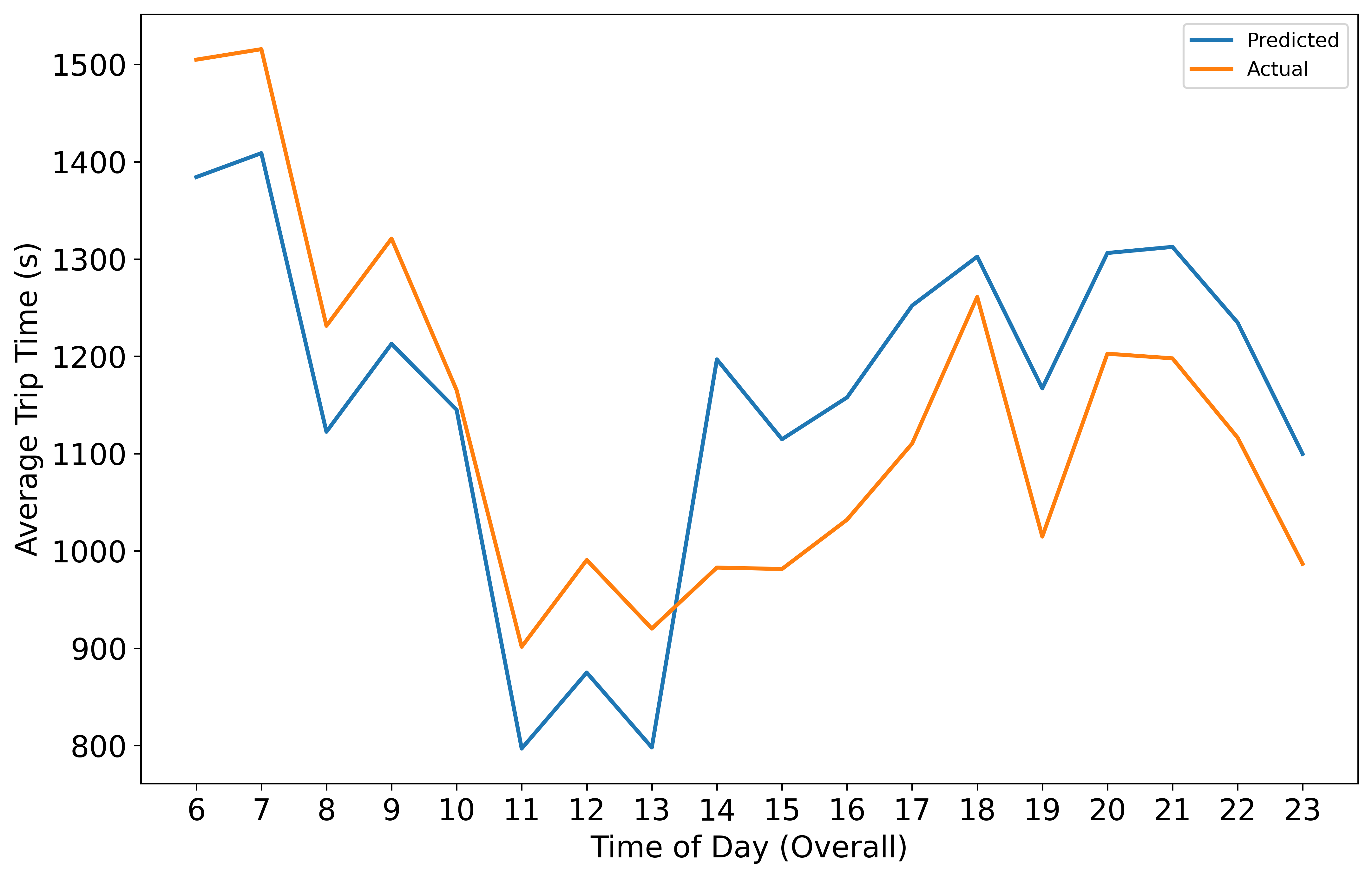}
        \label{fig11_subfigure4}
    }
    \subfigure[Faisal Avenue]{%
        \includegraphics[width=0.48\linewidth]{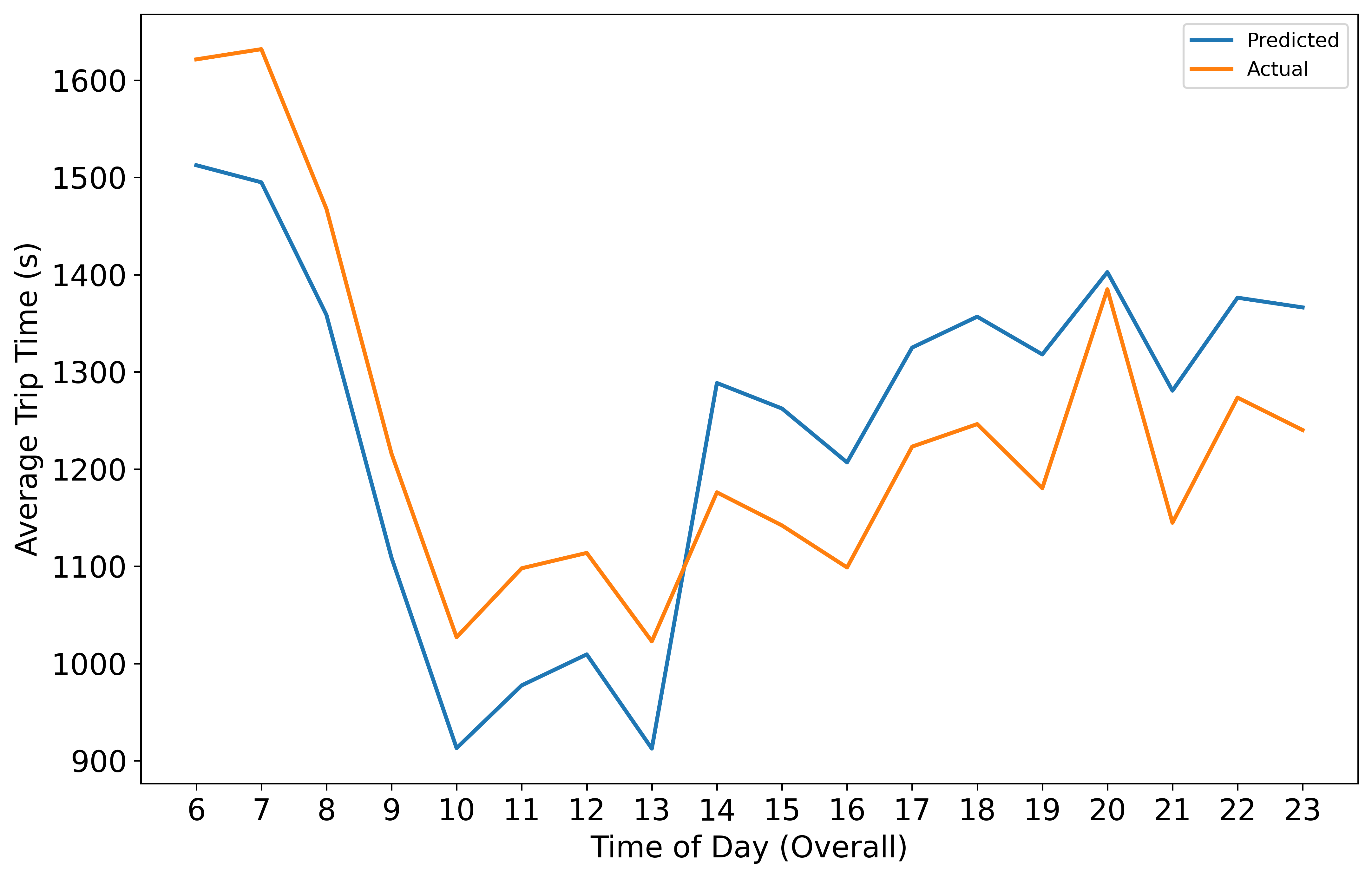}
        \label{fig11_subfigure5}
    }
    \hfill
    \subfigure[Agha Shahi Road]{%
        \includegraphics[width=0.48\linewidth]{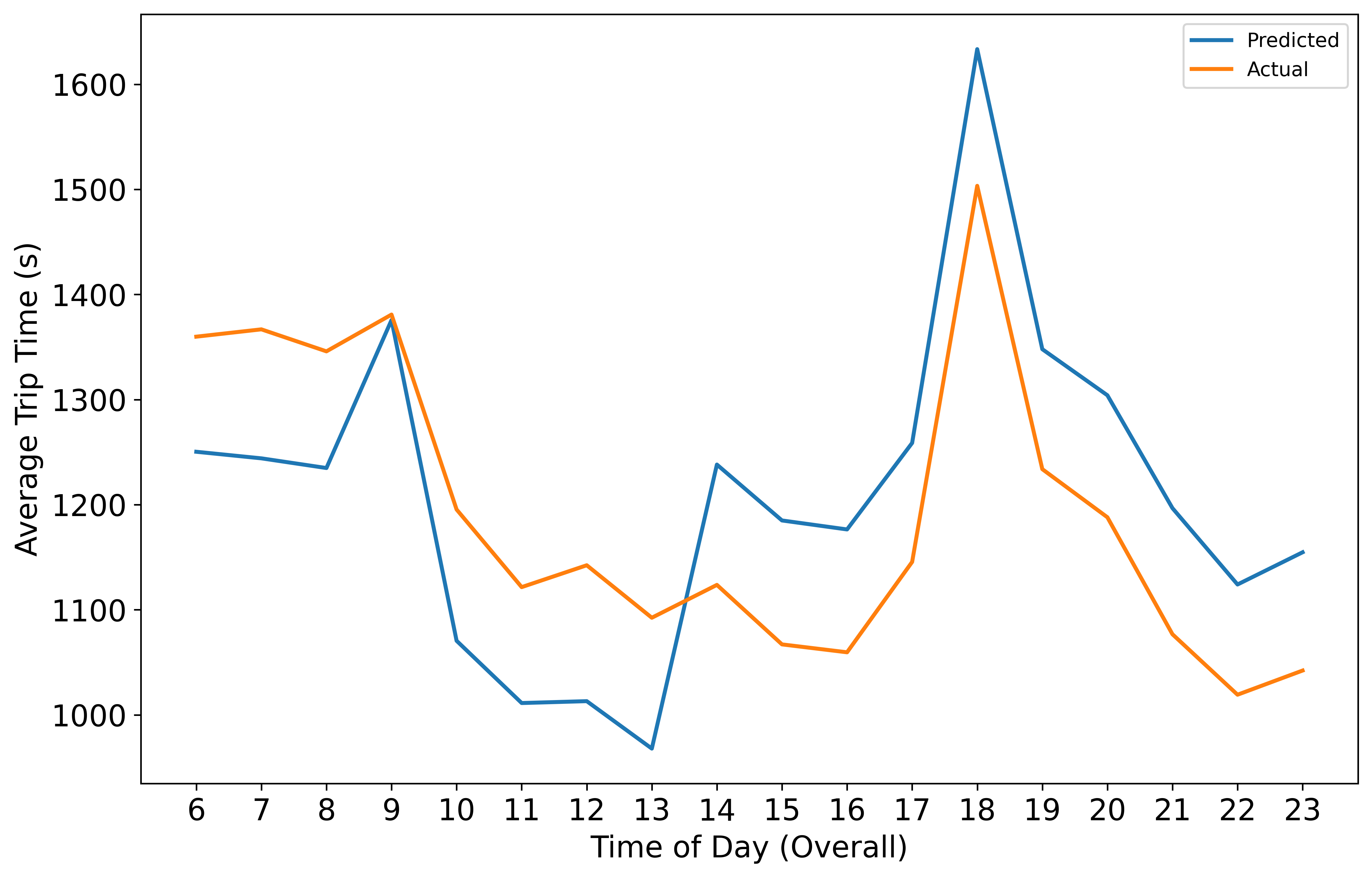}
        \label{fig011_subfigure6}
    }
    \caption{ANN Actual vs Predicted}
    \label{fig11}
\end{figure}

Determining these hyperparameters table \ref{tab3: parameters} includes rounds of experimentation, where various combinations are tested and their performance is evaluated using validation sets. The objective is to find the setups that produce precise predictions without overfitting the training data. The choice of ReLU as the activation function and batch is intended to exploit benefits, including computing efficiency, minimization of vanishing gradients, and introduction of non-linearity, in order to improve training and performance, with the ultimate goal of achieving accurate traffic time prediction. The root mean square error was between 1.5 and 3 minutes on each road, and 109 seconds on all roads. The graph of actual vs predicted values obtained with ANN on most frequent roads is shown in Figure \ref{fig11}.

\begin{table}
\centering
\caption{Travel Time Model Parameters}
\label{tbl:travel_time_models}
\begin{tabular}{|c|c|c|c|c|}
\hline
\textbf{Model} & \textbf{Neurons} & \textbf{Batch Size} & \textbf{Activation} & \textbf{Epocs} \\ \hline
ANN & 256 & 128 & ReLU & 200 \\ \hline
MLP & 64 \& 32 & 128 & ReLU & 200 \\ \hline
LSTM & 64 & 128 & ReLU & 50 \\ \hline
\end{tabular}
\label{tab3: parameters}
\end{table}

\begin{figure}[!htb]
    \centering
    \subfigure[Islamabad Expressway]{%
        \includegraphics[width=0.48\linewidth]{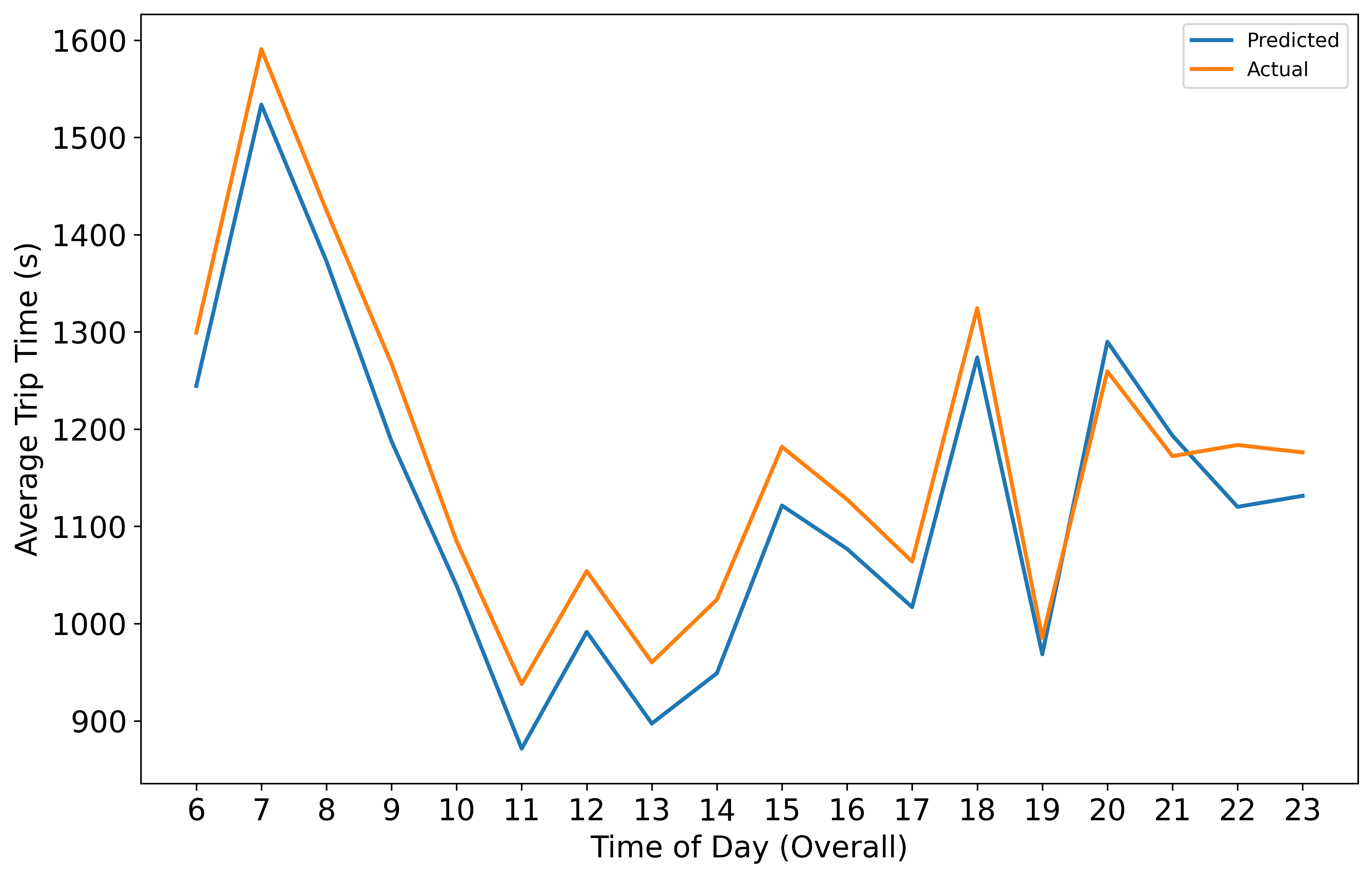}
        \label{fig12_subfigure1}
    }
    \hfill
    \subfigure[Srinagar Highway]{%
        \includegraphics[width=0.48\linewidth]{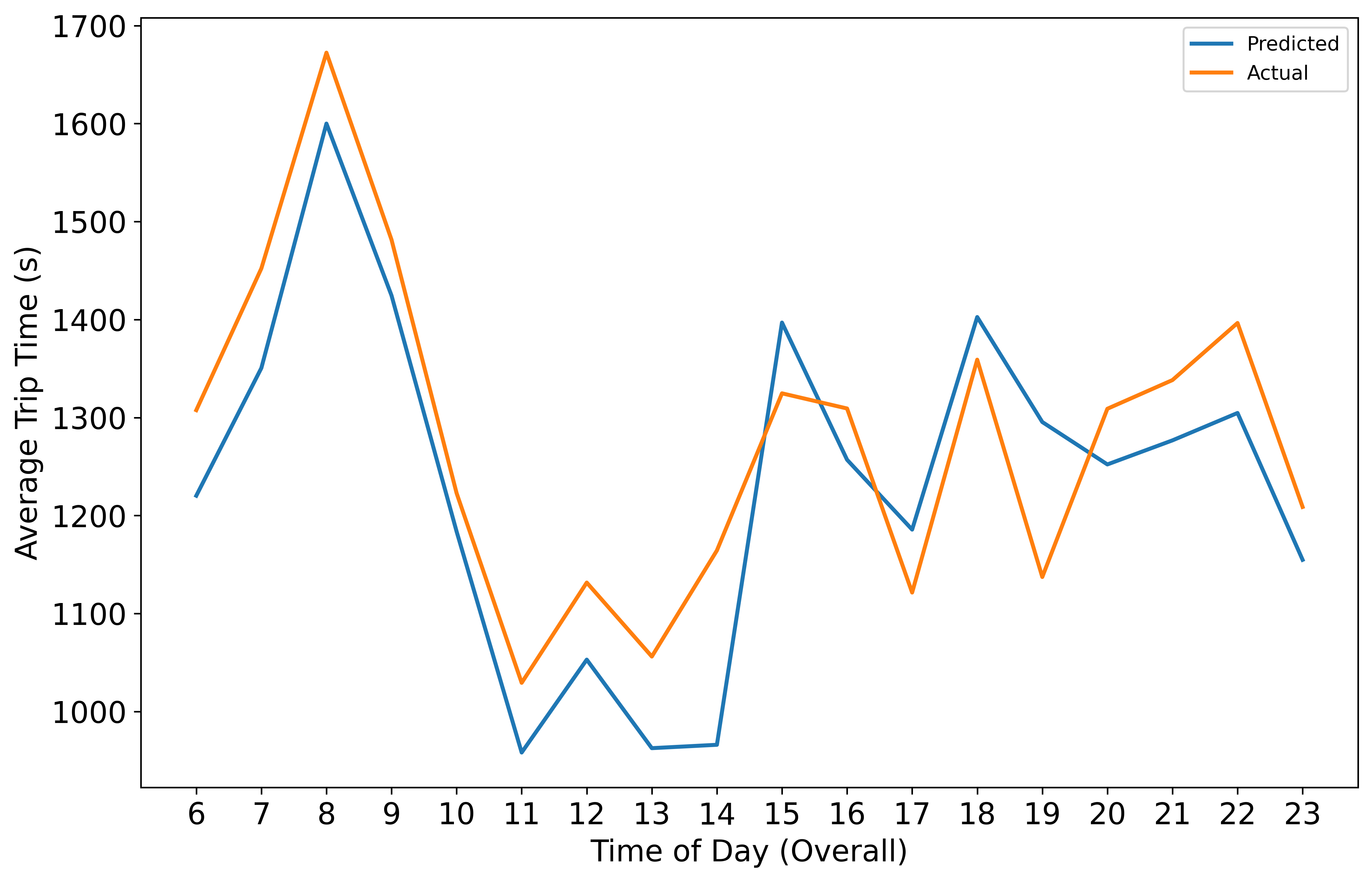}
        \label{fig12_subfigure2}
    }
    \subfigure[Khayaban-e-Iqbal]{%
        \includegraphics[width=0.48\linewidth]{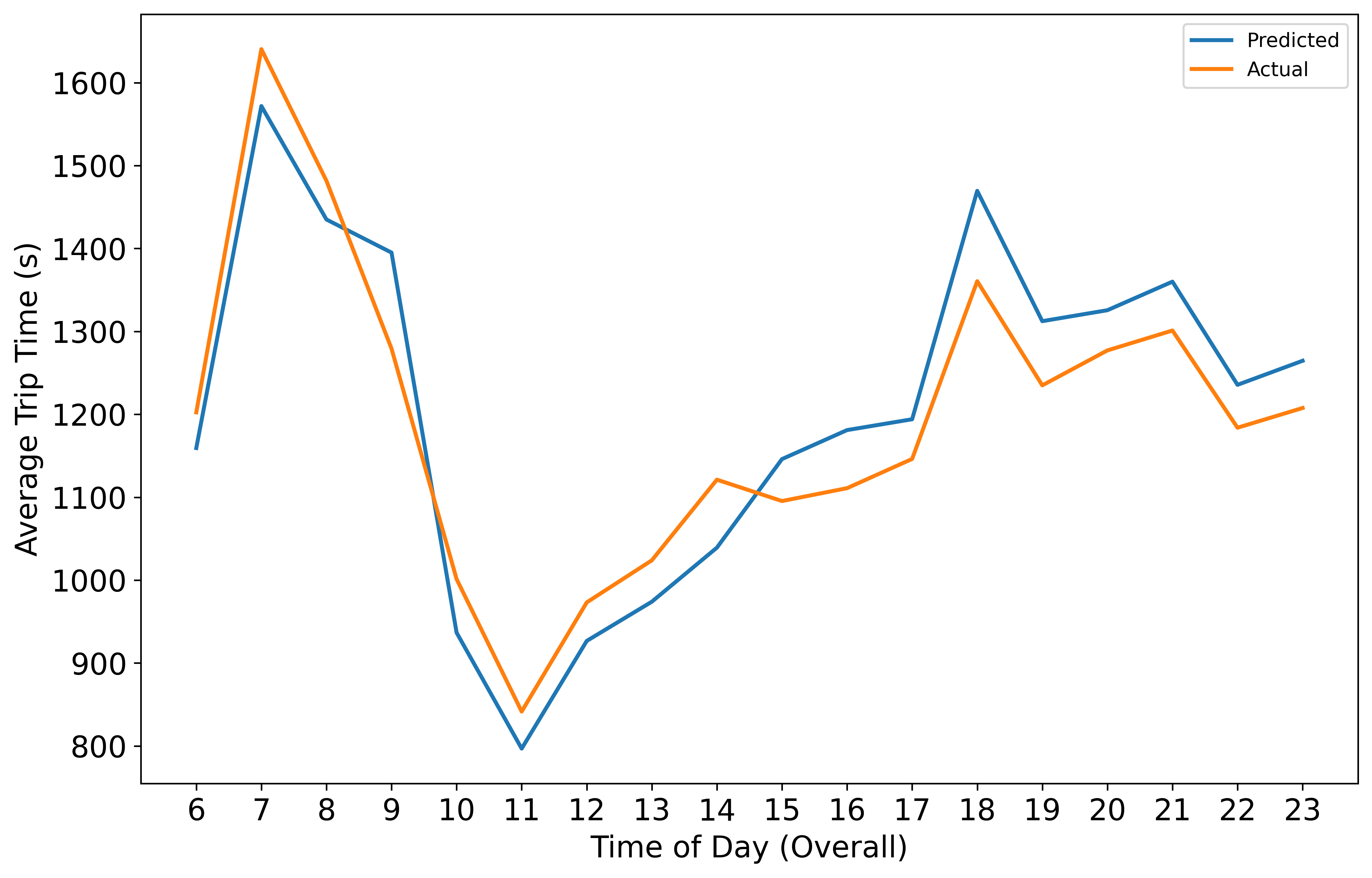}
        \label{fig12_subfigure3}
    }
    \hfill
    \subfigure[Jinnah Avenue]{%
        \includegraphics[width=0.48\linewidth]{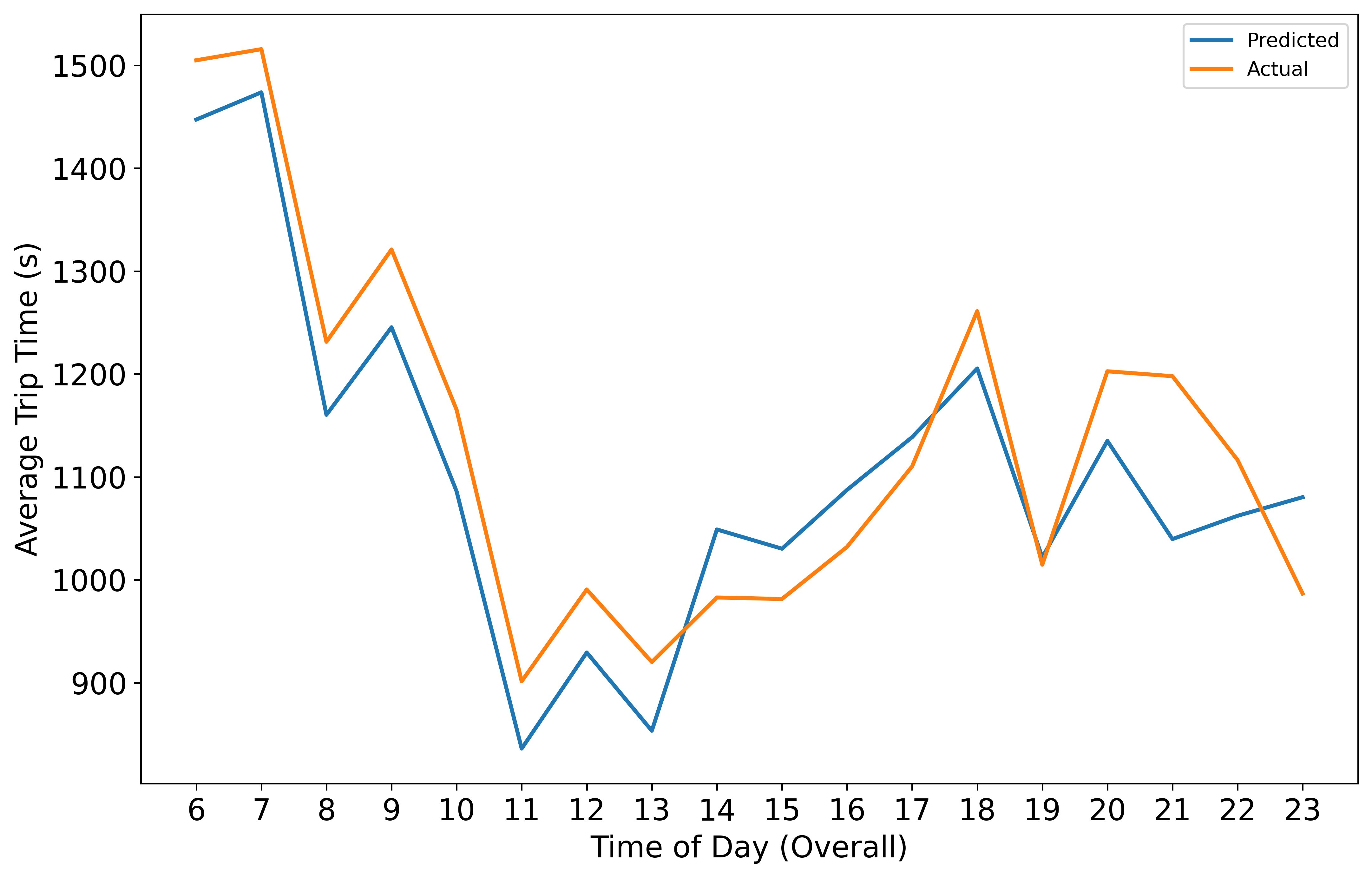}
        \label{fig12_subfigure4}
    }
    \subfigure[Faisal Avenue]{%
        \includegraphics[width=0.48\linewidth]{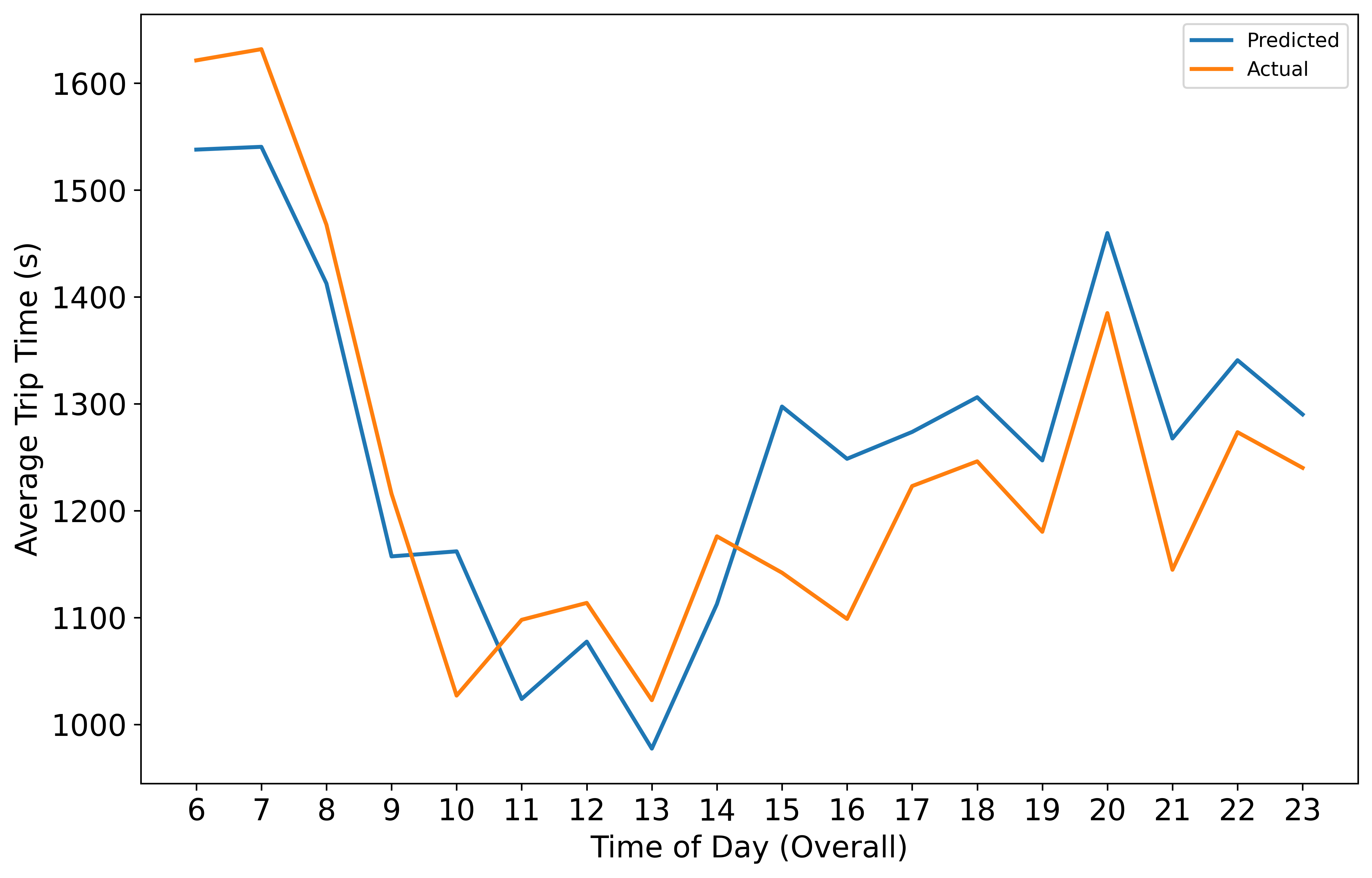}
        \label{fig12_subfigure5}
    }
    \hfill
    \subfigure[Agha Shahi Road]{%
        \includegraphics[width=0.48\linewidth]{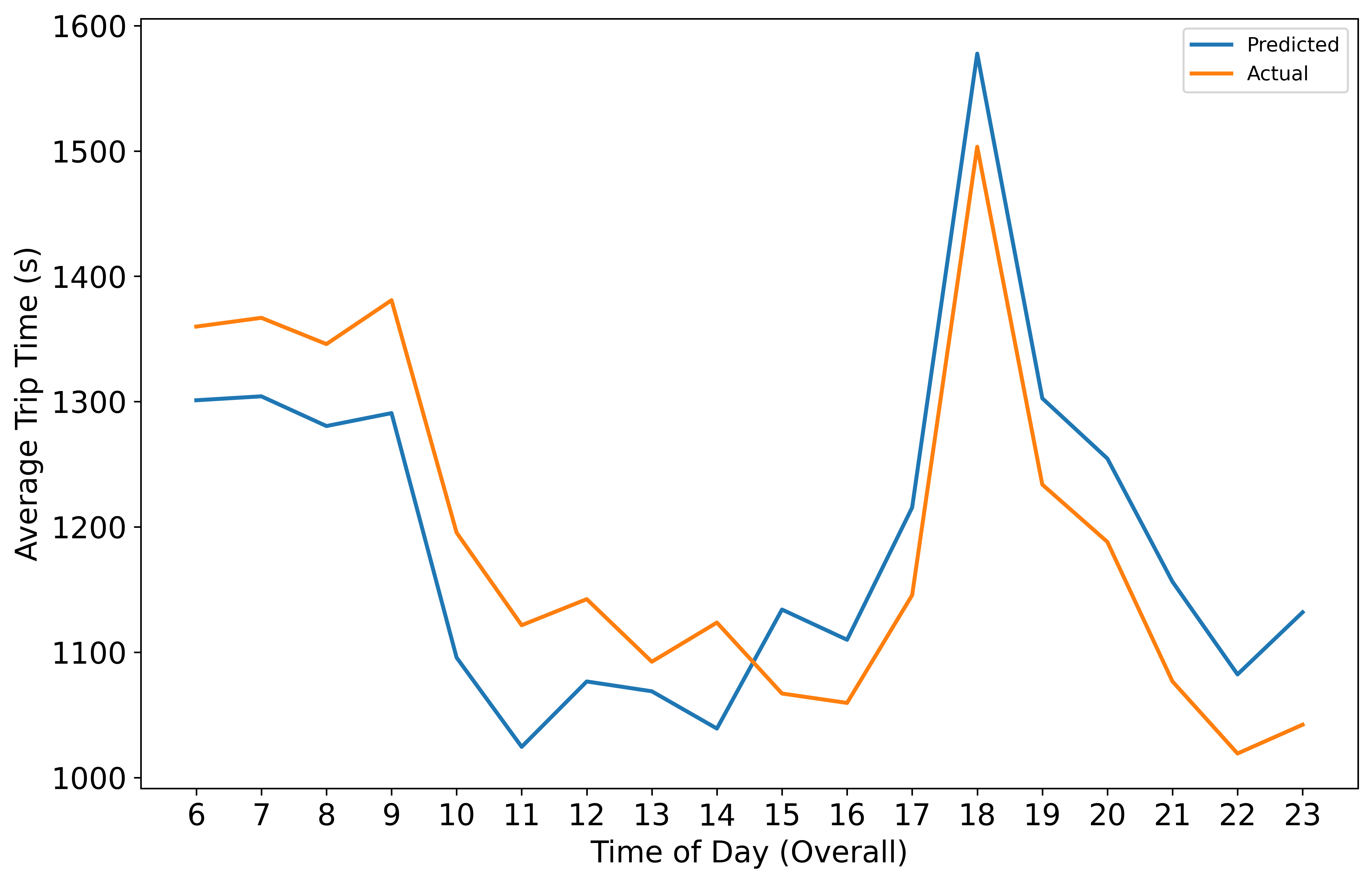}
        \label{fig012_subfigure6}
    }
    \caption{MLP Actual vs Predicted}
    \label{fig12}
\end{figure}

In addition to ANN, we employed MLP to forecast on the validation dataset after training. Each road had a root mean square error of half a minute (approximate), and all roads dataset had a root mean square error of 39 seconds. The graph of actual vs predicted values obtained with MLP on most frequent roads is shown in Figure \ref{fig12}.

\begin{figure}[!htb]
    \centering
    \subfigure[Islamabad Expressway]{%
        \includegraphics[width=0.48\linewidth]{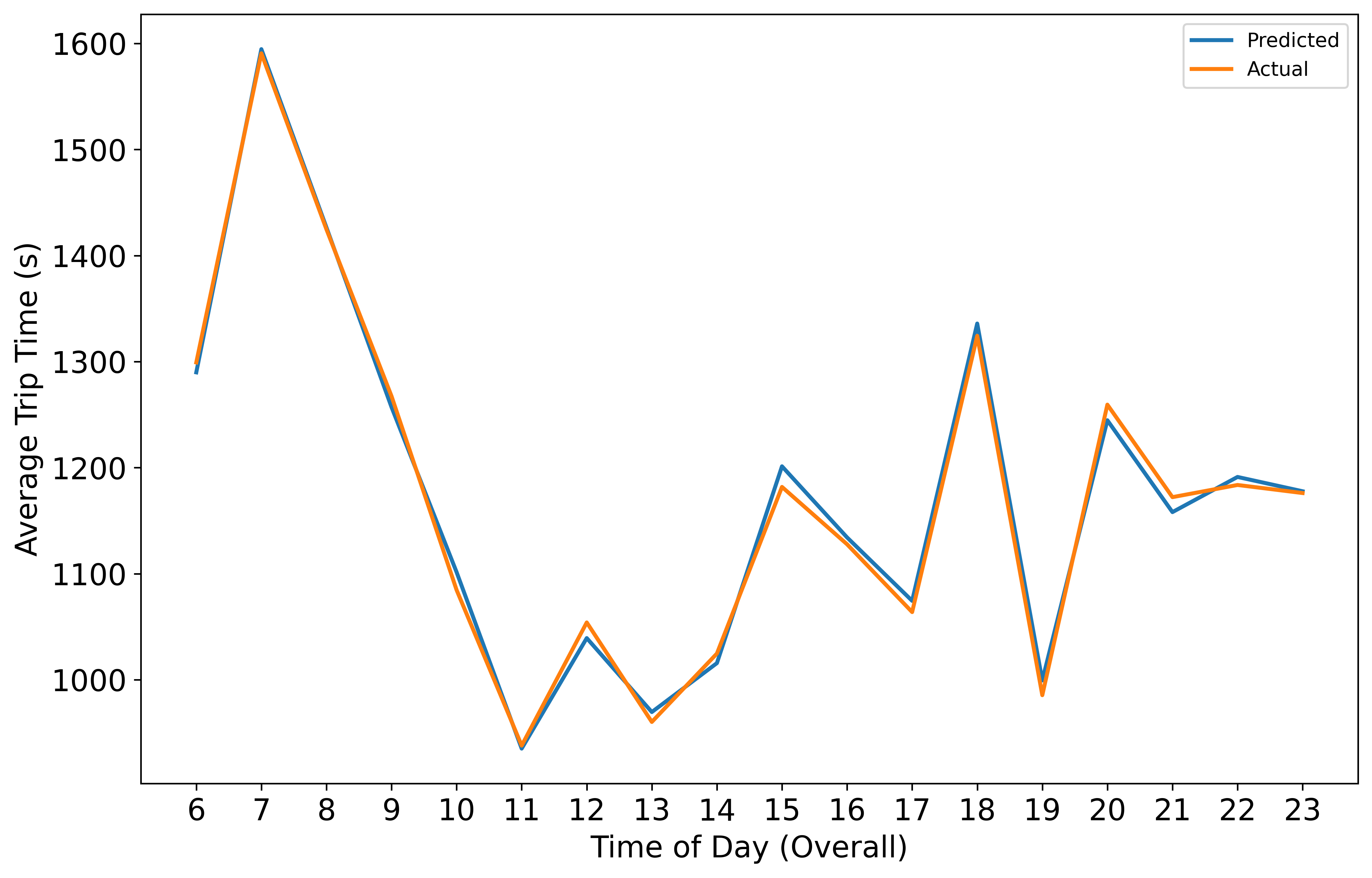}
        \label{fig13_subfigure1}
    }
    \hfill
    \subfigure[Srinagar Highway]{%
        \includegraphics[width=0.48\linewidth]{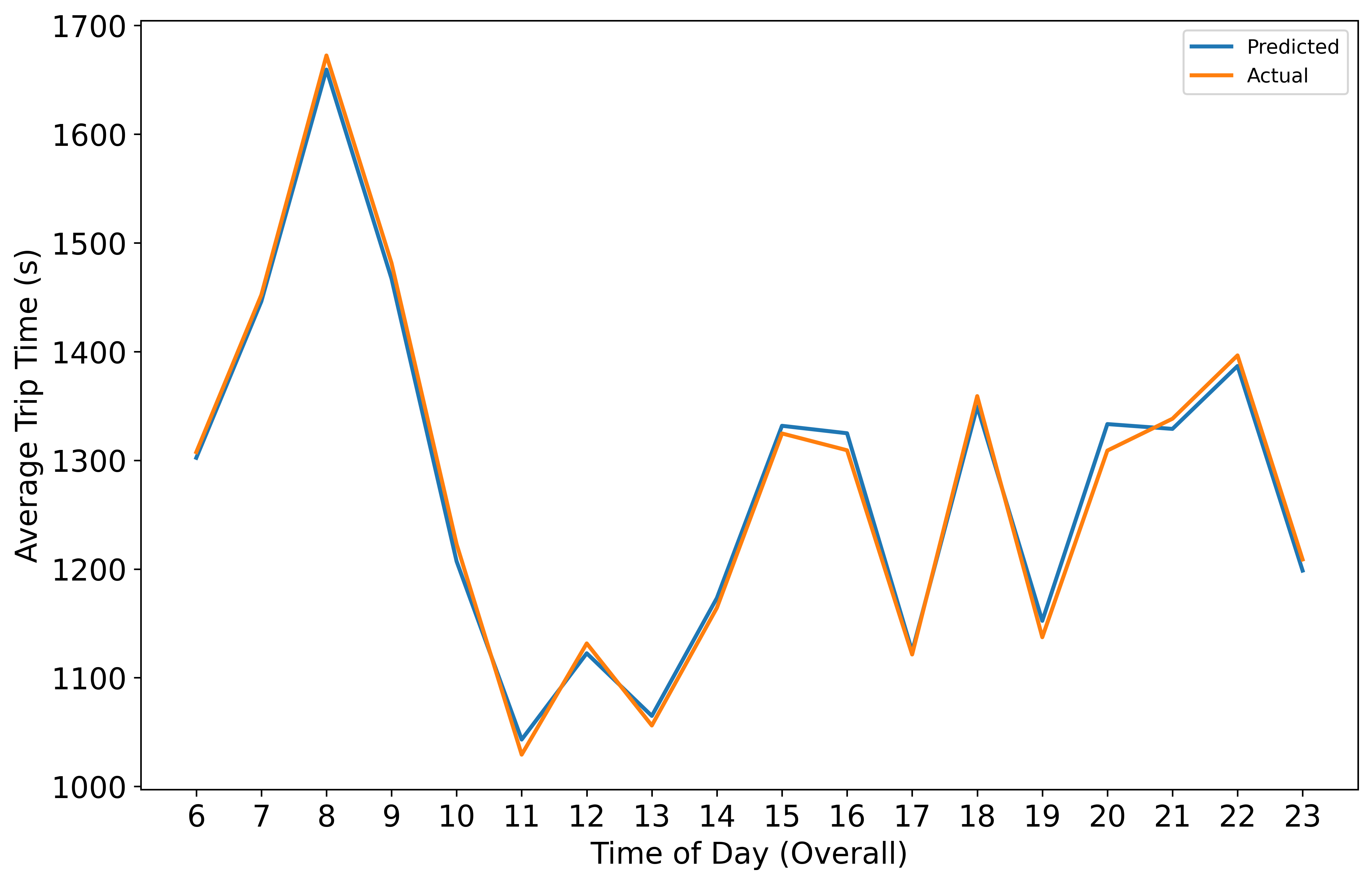}
        \label{fig13_subfigure2}
    }
    \subfigure[Khayaban-e-Iqbal]{%
        \includegraphics[width=0.48\linewidth]{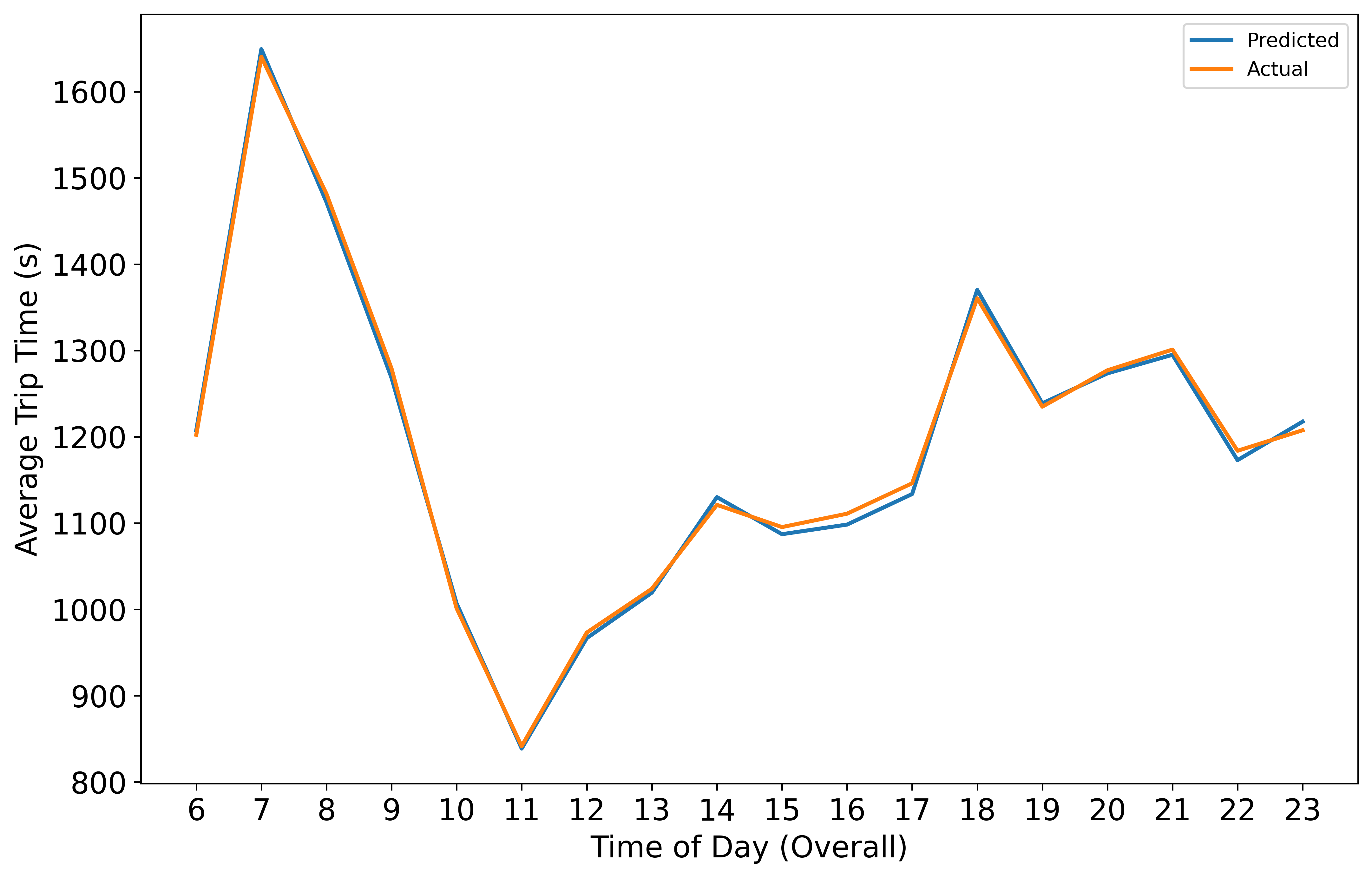}
        \label{fig13_subfigure3}
    }
    \hfill
    \subfigure[Jinnah Avenue]{%
        \includegraphics[width=0.48\linewidth]{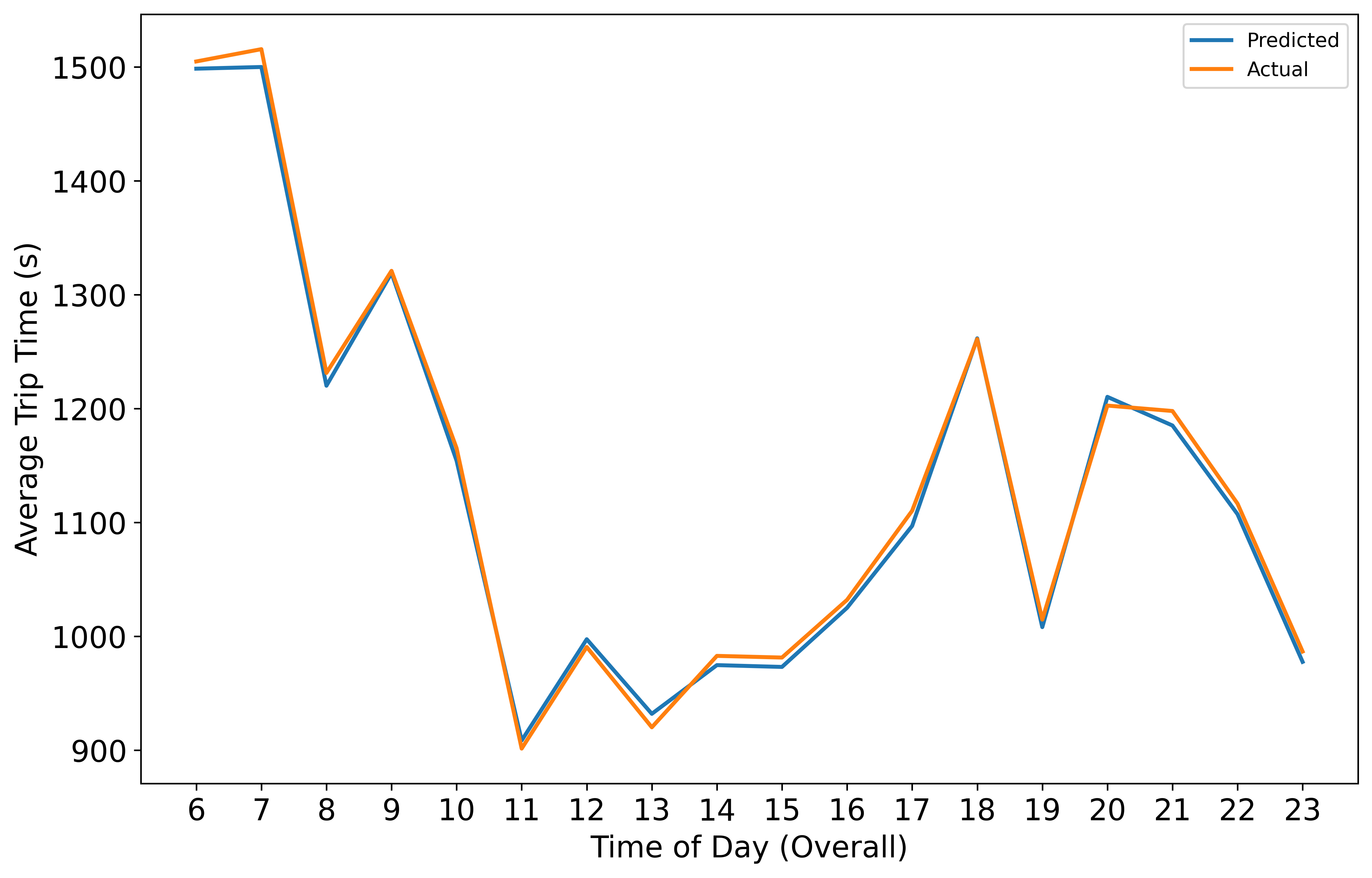}
        \label{fig13_subfigure4}
    }
    \subfigure[Faisal Avenue]{%
        \includegraphics[width=0.48\linewidth]{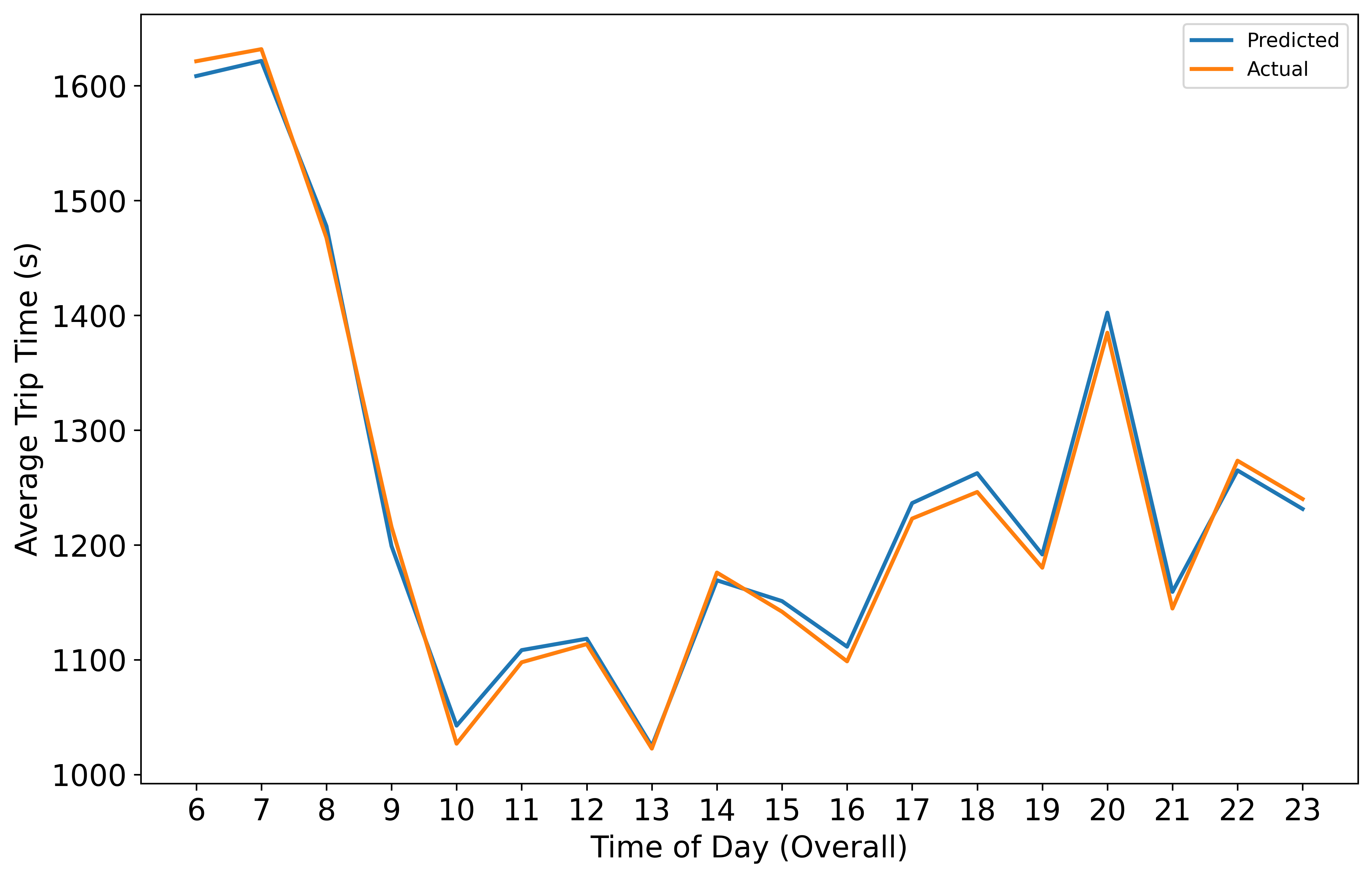}
        \label{fig13_subfigure5}
    }
    \hfill
    \subfigure[Agha Shahi Road]{%
        \includegraphics[width=0.48\linewidth]{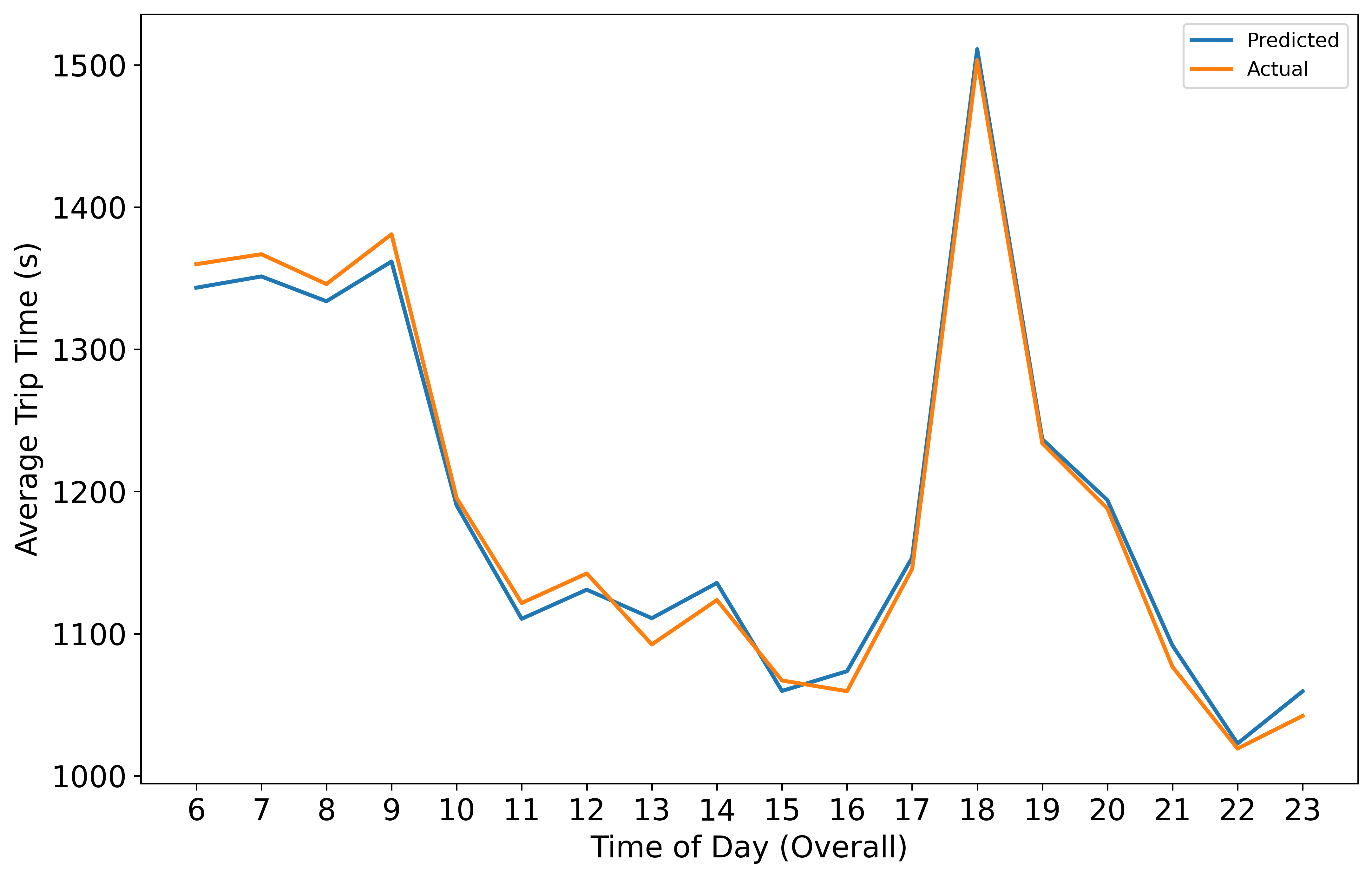}
        \label{fig013_subfigure6}
    }
    \caption{LSTM Actual vs Predicted}
    \label{fig13}
\end{figure}

Similarly, We trained an LSTM model on the same pre-processed data. Each road had a root mean square error of nearly 25 seconds, while the dataset for all roads had a root mean square error of 16.99 seconds. The graph of actual vs predicted values obtained using LSTM on most frequent roads is shown in Figure \ref{fig13}.

\begin{table}[htbp]
\caption{Performance Comparison of ANN, MLP, and LSTM}
\centering
\begin{tabular}{|p{2cm}|c|c|c|c|c|c|}
\hline
\multirow{2}{*}{\textbf{Road Names}} & \multicolumn{2}{c|}{\textbf{ANN}} & \multicolumn{2}{c|}{\textbf{MLP}} & \multicolumn{2}{c|}{\textbf{LSTM}} \\
\cline{2-7}
& \textbf{\textit{RMSE}} & \textbf{\textit{MAE}} & \textbf{\textit{RMSE}} & \textbf{\textit{MAE}} & \textbf{\textit{RMSE}} & \textbf{\textit{MAE}} \\
\hline
Islamabad Expressway & 143.53 & 43.35 & 48.47 & 18.28 & \textbf{24.28} & \textbf{12.80} \\
Srinagar Highway & 122.25 & 52.71 & 29.11 & 16.46 & \textbf{23.96} & \textbf{15.20} \\
Khayaban e Iqbal & 90.48 & 38.57 & 28.82 & 20.08 & \textbf{26.60} & \textbf{15.48} \\
Jinnah Avenue & 144.84 & 56.39 & 36.66 & 20.70 & \textbf{28.28} & \textbf{18.95} \\
Faisal Avenue & 131.21 & 59.25 & 43.15 & 22.56 & \textbf{32.86} & \textbf{18.68} \\
Agha Shahi Road & 157.12 & 57.68 & 51.32 & 22.28 & \textbf{21.10} & \textbf{12.90} \\
\hline
\end{tabular}
\label{tab2:results}
\end{table}

The three cutting-edge \cite{refcon1_6894591, refcon2_gobezie2020machine} methods were evaluated based on their performance in terms of RMSE and MAE on the six most often used routes are shown in Table \ref{tab2:results} . Observing the results, it's evident that the LSTM model regularly surpasses both ANN and MLP across all road segments in terms of both RMSE and MAE. LSTMs are highly effective in tasks that involve sequential data and long-term dependencies, making them especially suitable for predicting trip time. This is because they are able to understand the temporal dynamics and changing patterns in traffic data, which is essential in this context.

Due to their capacity to effectively store and update data over time intervals, they surpass shallower networks such as ANNs and MLPs in recognising the complexity of traffic patterns and generating more precise forecasts for potentially offering valuable insights for transportation planning and management.

\section*{Conclusion}
Trip Time Prediction (TTP) is a crucial element of an Intelligent Transportation System. Accurately predicting TTP results in substantial enhancements in overall transportation-related matters.  
In this study, We used Islamabad, Pakistan, GPS data obtained from sensors. We performed extensive computations to refine the dataset and experimented with numerous tools for Map Matching and trajectory simplification. Finally, after weeks of data enrichment using OSM and OSRM APIs on local servers, we obtained map-matched trajectories. We identified six most frequent routes for this study. On these selected routes, we trained three state-of-the-art approaches namely, a shallow ANN, a deep MLP and an LSTM for TTP. We obtained significantly improved performance with LSTM as compared to ANN and MLP for trip time predictions due to the fact that LSTM is a specialized time-series model in contrast to ANN and MLP.

In the future, we will utilize our models to develop an application for Islamabad residents, as navigation companies have expressed interest in such solutions. Moreover, we also plan to extend this work by adding other exogenous features with traffic data.

\bibliographystyle{ieeetr}
\bibliography{conference_101719.bib}
\vspace{12pt}

\end{document}